\documentclass{article}

\usepackage[T1]{fontenc}
\usepackage[utf8]{inputenc}

\usepackage{microtype}

\usepackage{inconsolata}

\usepackage{enumitem}
\usepackage{graphicx}
\usepackage{caption} 
\usepackage{booktabs}
\usepackage{float}
\usepackage{diagbox}
\usepackage{multirow}
\usepackage{subcaption}
\usepackage{amssymb}
\usepackage{bbold}
\usepackage{makecell}   
\usepackage{siunitx}   
\usepackage{tabularx}
\usepackage{times}
\usepackage{amsmath}
\usepackage{amsfonts}
\usepackage{amsthm}
\usepackage{cleveref}
\usepackage{float}
\usepackage{wrapfig}
\usepackage[ruled,vlined]{algorithm2e}
\usepackage{placeins}
\usepackage[numbers]{natbib}
\usepackage[table]{xcolor}
\usepackage{multirow}
\usepackage{diagbox}
\usepackage{tikz}
\usepackage[a4paper,scale=0.85]{geometry}
\usepackage{array}

\newcolumntype{C}{>{\centering\arraybackslash}X}

\newtheorem{example}{Example}

\newcommand{\TriCell}[2]{%
  \begin{tikzpicture}[baseline=(current bounding box.center),
                      every node/.style={font=\small}]
    \draw (0,0) rectangle (1.4,1.4);
    \draw (0,1.4) -- (1.4,0);
    \fill[gray!30] (0,1.4) -- (1.4,1.4) -- (1.4,0) -- cycle;
    \node at (1.05,1.05) {#1};  
    \node at (0.35,0.35) {#2};   
  \end{tikzpicture}%
}

\title{CHBench: A Cognitive Hierarchy Benchmark for Evaluating Strategic Reasoning Capability of LLMs}
\date{}

\author{%
Hongtao Liu\thanks{Equal Contributions}, Zhicheng Du$^*$, Zihe Wang, Weiran Shen\thanks{Corresponding Author} \\
 \texttt{\{ht6,duzhicheng,wang.zihe,shenweiran\}@ruc.edu.cn} \\
 Gaoling School of Artificial Intelligence\\ Renmin University of China
}

\begin{document}

\maketitle

\begin{abstract}

Game-playing ability serves as an indicator for evaluating the strategic reasoning capability of large language models (LLMs).
While most existing studies rely on utility performance metrics, 
which are not robust enough due to variations in opponent behavior and game structure.
To address this limitation, we propose \textbf{Cognitive Hierarchy Benchmark (CHBench)}, a novel evaluation framework inspired by the cognitive hierarchy models from behavioral economics.
We hypothesize that agents have bounded rationality --- different agents behave at varying reasoning depths/levels.
We evaluate LLMs' strategic reasoning through a three-phase systematic framework, utilizing behavioral data from six state-of-the-art LLMs across fifteen carefully selected normal-form games.
Experiments show that LLMs exhibit consistent strategic reasoning levels across diverse opponents, confirming the framework's robustness and generalization capability.
We also analyze the effects of two key mechanisms (Chat Mechanism and Memory Mechanism) on strategic reasoning performance. 
Results indicate that the Chat Mechanism significantly degrades strategic reasoning, whereas the Memory Mechanism enhances it.
These insights position CHBench as a promising tool for evaluating LLM capabilities, with significant potential for future research and practical applications.

\label{sec:abstract}
\end{abstract}

\section{Introduction}
\label{sec:intro}

The strategic reasoning ability of large language models (LLMs) has attracted much research attention. Such ability not only helps us understand how intelligent LLMs are, but also enables sophisticated applications such as realistic human behavior simulations and autonomous decision-making (e.g., interactive simulacra of human behavior \citep{akata2023playing,park2023generative}, constituting an essential pathway toward Artificial General Intelligence (AGI). Although recent years have seen huge improvements in the strategic reasoning ability of LLMs, how to evaluate such ability of LLMs quantitatively remains a critical challenge.

Game theory \citep{fudenberg1991game,myerson2013game,owen2013game,von1944theory} is a field of research that studies strategic reasoning and interactions among rational decision-makers. The mathematical framework of game theory helps to model key aspects in intelligent decision-making, such as modeling opponent behaviors and long-term planning under uncertain and competitive environments. Thus, the ability to play games could serve as a measure for evaluating the strategic reasoning capability of LLMs. Indeed, many existing studies focus on game-theoretic reasoning abilities of LLMs, especially in the setting of normal-form games or repeated games due to their relatively simple representations and the connection to classic game theory \citep{akata2023playing,costarelli2024gamebench,lore2024strategic,mao2025alympics,ross2024llm}.

\begin{figure*}[!t]
    \centering
    \includegraphics[width=1\linewidth]{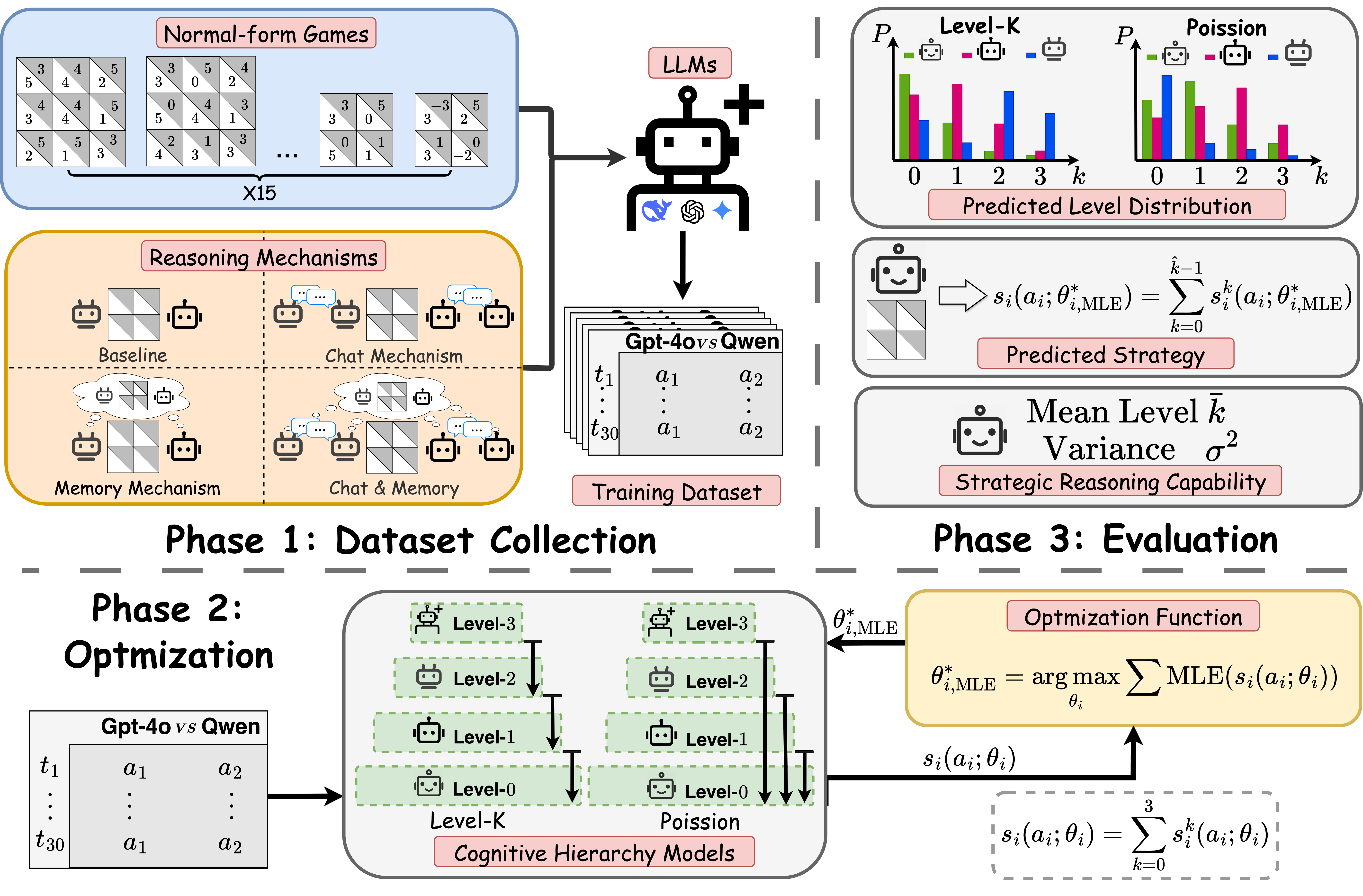}
    \caption{The full architecture of CHBench framework.}
    \label{fig:framework}
\end{figure*}

Most existing studies focus on the utility performance of the LLMs across different normal-form games to evaluate the strategic reasoning capability \citep{duan2024gtbench,feng2024survey,guo2024economics,hu2024survey,zhang2024llm}. 
This approach is reasonable in its own right since game players always aim to maximize their utilities. However, the robustness of the method is limited due to both the game utility structure and uncertainties in the opponent's behaviors \citep{goeree2001ten,stahl1995players,stirling2003satisficing,wright2010beyond,wright2017predicting}. 
Here we provide a symmetric game to show the limit of this approach.
\begin{table}[!htb]
\caption{A symmetric two-player game}
\centering
\renewcommand\arraystretch{0.8}   % 行距
\setlength{\tabcolsep}{0 pt}       % 列间距
\scalebox{0.8}{
\begin{tabular}{c cc}   % 只有数据列之间有竖线
      & \textbf{A} & \textbf{B} \\ %\hline
\textbf{A~~}   & \TriCell{-8}{-8} & \TriCell{0}{5} \\ %\hline
\textbf{B~~}   & \TriCell{5}{0} & \TriCell{1}{1} \\
\end{tabular}}
\label{tab:exp_utility_fail}
\end{table}

\begin{example}
Consider the game shown in Table \ref{tab:exp_utility_fail}. There are two players in the game (the row player and the column player), both with two available actions (action A and action B). The two numbers in each cell represent the utilities of the row player and the column player, respectively.

It is easy to check that the Nash equilibrium of the game is for both players to choose action A with probability $\frac{1}{3}$ and action B with probability $\frac{2}{3}$.  Suppose that the row player is na\"ive and chooses both actions uniformly randomly, while the column player is able to compute the Nash equilibrium and uses the equilibrium strategy. In this case, the expected utilities of the row player and the column player are $\frac{2}{3}$ and $-\frac{1}{6}$. Even if the column player is able to accurately predict the row player's strategy and best responds accordingly (which is to choose action B with probability 1), the expected utility of the row player is 3, while the column player's expected utility is only 0.5.

In either case, the column player has stronger strategic reasoning capability, but still obtains a much lower expected utility than the row player.
\end{example}

To address this limitation, we propose the Cognitive Hierarchy Benchmark (CHBench) framework to evaluate the strategic reasoning capability of LLMs. The cognitive hierarchy model \citep{binmore1987modeling} and its level-$k$ extension originate in behavioral economics, and are used to describe human thought processes.
To the best of our knowledge, this paper is the first work that adopts such models to evaluate the strategic reasoning capability of LLMs.
Unlike standard game theory (which assumes perfect rationality), the cognitive hierarchy model considers limited human rationality -- different people think at different levels of strategic depth. This model focuses on the players' ability to model and predict the behaviors of their opponents rather than the obtained utility, and thus aligns better with our goal to measure the reasoning capability of LLMs.

In the CHBench framework, each pure strategic player performs \textit{iterative best response} to the strategies of others.
For instance, consider a two-player game, player $1$ at level $k$ adopts the best response strategy against player $2$'s strategies at levels below $k$.
Specifically, we examine two distinct CH models: the \textit{Level-K CH} model and the \textit{Poisson CH} model. 
The main difference between the two models lies in their behavioral assumptions: the Level-K CH model assumes that each level‑$k$ player performs iterative best response to level-$(k-1)$ players (\textit{myopic iterative best response}), whereas the Poisson CH model assumes that each player performs iterative best response to all players at below level-$k$ (\textit{omniscient iterative best response}).

Figure \ref{fig:framework} illustrates the architecture of CHBench. The implementation of CHBench framework contains three phases: the data collection phase, the optimization phase, and the evaluation phase. 
During the former two phases, we collect behavioral data from LLMs across various game matrices and optimize the model parameters using the method of Maximum Likelihood Estimation (MLE). 
In the evaluation phase, given any normal-form game and the specified parameters, the model generates the level distribution for each agent, and the predicted strategy at each level for each agent.
The mean of the level distribution represents the agent's average strategic reasoning capability in the game, while the variance indicates the stability of the agent's strategic behaviors. \Cref{sec:chbench} introduces the process of three phases.

We evaluate the strategic reasoning capability of six LLMs across different games and mechanisms. Our key findings are briefly summarized below.
First, we show that setting the maximum level of cognitive hierarchy framework at $\hat{k}=4$ is appropriate as increasing the maximum level yields negligible improvements in likelihood while amplifying strategic instability.
Second, we observe that the same LLM exhibit only slight fluctuation in expected strategies when facing different opponents in the same game. Confirming that our framework effectively measures the consistency of the strategic reasoning capability of LLMs.
Finally, we propose two mechanisms to explore their effect on the strategic reasoning capability of LLMs: \textbf{Chat Mechanism} and \textbf{Memory Mechanism}. 
Experiments show that the Chat Mechanism significantly degrades strategic reasoning ability, while the Memory Mechanism enhances it. 
When combining both mechanisms, the resulting strategic reasoning capability fall between their individual effects.

Our main contributions can be summarized as follows:
\begin{itemize}[leftmargin=*]
    \item We introduce the CHBench framework based on Cognitive Hierarchy models from behavioral economics to evaluate the strategic reasoning capability of LLMs.
    \item We demonstrate the effectiveness of CHBench in evaluating strategic reasoning capability of LLMs by showing that the consistency of strategy when facing different opponents in the same game.
    \item We examine the impact of the Chat Mechanism and Memory Mechanism on LLMs' strategic reasoning capability using the CHBench framework.
\end{itemize}

\section{Cognitive Hierarchy Model}
\label{sec:prelim}

We model each agent's strategic reasoning capability as a linear combination of up to $\hat{k}$ cognitive levels (we justify our choice of $\hat{k}=4$ in \Cref{subsec:k_hat}). 
The key assumption is that each agent performs \textit{iterative best response} to the opponent's strategy -- specifically, a level-$k$ agent \textit{best responds} to opponents at levels below $k$.
Here, \textit{best response}, a fundamental concept in game theory, refers to the strategy that maximizes expected payoff given the opponent's strategy.

In this paper, we examine two variants of CH models: the \textit{Level-K CH} model and the \textit{Poisson CH} model. 
First, we introduce necessary notation. 
Consider a two-player normal-form game where each agent $i \in \{1,2\}$ has a finite action space $A_i$. For an agent $i$ with cognitive level $k \in \{0,\ldots,\hat{k}-1\}$, their conditional predicted strategy is denoted as $s_i^k \in \Delta(A_i)$. 
Both models contain parameters that require optimization. This section focuses on applying the framework to predict strategic capability with fixed parameters; we defer the optimization details to \Cref{subsec:optimization}. 

\subsection{Level-K CH Model}

We fix model parameters $\{\epsilon_1^k,\alpha_1^k,\epsilon_2^k,\alpha_2^k\}_{k\in\{0,\dots,\hat{k}-1\}}$, where $\epsilon_i^k$ denotes the error rate of agent $i$ with level $k$ and $\alpha_i^k$ the weight of level $k$ in agent $i$'s overall strategy.

Agents employ \textit{myopic iterative best response}: a level-$k$ agent best responds only to opponents at level $k-1$. 
We denote by $\text{BR}_{i}^{k}$ the set of agent $i$'s level-$k$ actions that best responds to the opponent's level-$(k-1)$ strategy. 
When level $k=0$, agents select actions uniformly at random from $A_i$, that is
$
    s_i^0(a_i)=1/|A_i|
$ for each action $a_i\in A_i$.
When level $k>0$, we introduce strategic uncertainty by assuming agent $i$ makes mistakes with probability $\epsilon_i^k$. 
Specifically, with probability $1-\epsilon_i^k$, the agent selects an action uniformly from $\text{BR}_{i}^{k}$; and with probability $\epsilon_i^k$, the agent chooses uniformly from $A_i - \text{BR}_i^k$.
Formally, for any $k>0$:
\begin{align*}
	s_i^k(a_i)&=
	\begin{cases}
		(1-\epsilon_k)/|\text{BR}_i^k|   &\text{if~} a_i \in \text{BR}_i^k~,  \\
		\epsilon_k/(|A_i-\text{BR}_i^k|) &\text{otherwise}~. 
	\end{cases}\quad \forall a_i\in A_i~.
	\label{eq:lk_s_i_k}
\end{align*}

The overall predicted strategy of agent $i$ is a linear combination of each level's predicted strategy:
\begin{align*}
	s_i(a_i)=\sum_{k=0}^{\hat{k}-1}\alpha_i^k\cdot s_i^k(a_i)~,\quad \forall a_i\in A_i~.
	% \label{eq:lk_s_i}
\end{align*}

\subsection{Poisson CH Model}
We fix model parameters $\lambda_1$ and $\lambda_2$, where $\lambda_i$ determines agent $i$'s level distribution (Poisson rate).

Agents employ \textit{omniscient iterative best response}: a level-$k$ agent best responds to opponents' strategies from all lower levels ($0$ to $k-1$). 
We denote by $\text{BR}_{i}^{0:k}$ the set of agent $i$'s level-$k$ best responses to this combination of opponent strategies.
Additionally, the Poisson CH model assumes that each agent $i$'s level follow a Poisson distribution with parameter $\lambda_i$.
Thus, in the view of level-$k$ agent $i$, his opponent’s strategy he best-responds to, $s_{-i}^k$, is a weighted combination of each level's predicted strategy from $0$ to $k-1$ according to a Poisson distribution. That is, for each action $a_{-i}\in A_{-i}$:
\begin{align*}
    s_{-i}^k(a_{-i})=\sum_{l=0}^{k-1}\frac{f_i(l)}{\sum_{l'=0}^{k-1}f_i(l')}\cdot s_{-i}^l(a_{-i})~,
\end{align*}
where $f_i(k) = \text{Poisson}(k|\lambda_i) = e^{-\lambda_i} \lambda_i/k!$.

When $k=0$, agent $i$ chooses an action from $A_i$ uniformly at random, that is
$
    s_i^0(a_i)=1/|A_i|
$ for $\forall  a_i\in A_i$.
When $k>0$, the conditional predicted strategy is as follows: 
	\begin{align*}
		s_i^k(a_i)&=
		\begin{cases}
			1/|\text{BR}_{i}^{0:k}| &	\text{if~} a_i \in \text{BR}_{i}^{0:k}~,\\
			0	&	 \text{otherwise}~.
		\end{cases}\quad \forall a_i\in A_i~.
	\end{align*}
The overall predicted strategy is a linear combination of each level's predicted strategy:
\begin{align*}
	s_i(a_i)=\sum_{k=0}^{\hat{k}-1} \frac{f_i(k)}{\sum_{l=0}^{k-1}f_i(l)}\cdot s_i^{k}(a_i)~,\quad\forall a_i\in A_i~.
\end{align*}

%\section{CHBench Framework}
\section{CHBench Framework}
\label{sec:chbench}

\Cref{fig:framework} illustrates the three phases of CHBench framework, in this section, we will give a detailed introduction of the construction and function of each component in the three phases.

\subsection{Dataset Collection}
\subsubsection*{LLMs} 
We use six LLMs to construct the training dataset. Three open-source LLMs: DeepSeek-chat-V3 \citep{bi2024deepseek}, Llama-3.1-70B-Instruct \citep{grattafiori2024llama}, and Qwen-max \citep{yang2024qwen2}. 
Three widely-used proprietary LLMs: GPT-4o, GPT-4o-mini \citep{achiam2023gpt}, and Gemini-1.5 Pro \citep{team2024gemini}.

\subsubsection*{Prompts}
We use four prompt templates: \textit{baseline prompt}, \textit{chat prompt}, \textit{memory prompt}, and \textit{chat \& memory prompt}. 
In each template, the game rules are explicitly elaborated, and the possible payoffs for every action profile are listed based on the specific game matrix. 
Building on the baseline prompt, the memory prompt and chat \& memory prompt further integrate historical information, while the chat prompt and chat \& memory prompt incorporate conversational data. The full prompts are provided in \Cref{sec:prompt_templates}.

\subsubsection*{Reasoning Mechanisms}
\label{subsec:mechanism}
We design three different mechanisms to affect LLMs' strategic reasoning capability.
\begin{itemize}
    \item \textbf{Chat Mechanism} 
    Before decision-making in any repeated game, both players are permitted to engage in two rounds of structured communication with their respective advisors (LLMs of the same type). 
    Each round consists of a question-and-answer exchange, allowing players to clarify and discuss the game rules in detail, and strategize and refine their decision-making approaches.
    \item \textbf{Memory Mechanism}
    Before each agent decides in subsequent rounds, we incorporate historical game information as memory into the prompt, including the round number, both agents' actions, and the resulting payoffs from previous rounds. 
    To examine how memory completeness influences strategic reasoning ability, we implement two memory mechanisms: Partial Memory, which retains only the $10$ most recent rounds of information, and Full Memory, which preserves all historical records from every previous round. 
    This design allows us to systematically compare the effects of limited versus complete historical information on agents' decision-making processes.
    \item \textbf{Chat \& Memory Mechanism}
    In this integrated mechanism, we combine the Chat Mechanism with the Memory Mechanism. Specifically, we incorporate all historical records from previous rounds into the agent's memory, and every $10$ rounds, agents are granted the opportunity to engage in two structured communication sessions with their respective advisors (LLMs of the same type), each consisting of a question-and-answer exchange. During these sessions, agents can analyze past interactions with their advisors, review historical data, and refine their strategies—enabling them to adapt and counter their opponent's tactics more effectively.

\end{itemize}

\subsubsection*{Normal-form Games}
We use 4 classic normal-form games (Prisoner's Dilemma, Coordination Game, Hawk-Dove Game, and Snowdrift Game) and 11 randomly generated normal-form games to conduct our experiments. 
There is no information leak during pre-training because classic games represent a class of games rather than a specific instance. 
Even for games of the same type, behavioral results can vary significantly across different instances.
We consider normal-form games between two players, each with a finite action space.
Each game's payoff structure can be shown in a payoff matrix (see \Cref{tab:exp_game} for examples).
The remaining games are shown in \Cref{sec:game_matrix_all}.

\begin{table}[!htb]
\caption{Examples of normal-form games.}
\centering
\begin{subtable}{0.48\linewidth}
\caption*{\#$1$: $2\times2$ and symmetric}
\centering
\renewcommand\arraystretch{0.8}   % 行距
\setlength{\tabcolsep}{0pt}       % 列间距
\scalebox{0.8}{
\begin{tabular}{c cc}   
      & \textbf{A} & \textbf{B} \\ %\hline
\textbf{A~~}   & \TriCell{3}{3} & \TriCell{5}{0} \\ %\hline
\textbf{B~~}   & \TriCell{0}{5} & \TriCell{1}{1} \\
\end{tabular}}
\label{tab:game_matrix_1_main}
\end{subtable}
\hfill
\begin{subtable}{0.48\linewidth}
\caption*{\#$15$: $3\times3$ and asymmetric}
\centering
\renewcommand\arraystretch{0.8}   % 行距
\setlength{\tabcolsep}{0pt}       % 列间距
\scalebox{0.6}{
\begin{tabular}{cc cc}   
      & \textbf{A} & \textbf{B} &\textbf{C}\\ %\hline
\textbf{A~~}   & \TriCell{3}{5}	& \TriCell{4}{4} & \TriCell{5}{2} \\ %\hline
\textbf{B~~}   & \TriCell{4}{3}	& \TriCell{4}{4} & \TriCell{5}{1} \\
\textbf{C~~}   & \TriCell{5}{2}	& \TriCell{5}{1} & \TriCell{3}{3} \\
\end{tabular}}
\label{tab:game_matrix_15_main}
\end{subtable}
\label{tab:exp_game}
\end{table}

\subsubsection*{Training Dataset} 
For each experiment, we select a specific normal-form game, assign two LLMs as the row and column players, and endow a reasoning mechanism for both LLMs.
To collect the behavioral data, we repeatedly query both LLMs over $T=30$ rounds, and record their action choices in each round.
Notice that if the game matrix is asymmetric, that is, the payoff structure faced by the \textit{row player} and the \textit{column player} differs, then we need to swap the roles of both LLMs and conduct one more experiment. 
Given each pair of LLMs, we gather $600$ rounds of action choice data under $15$ different normal-form games.
By iterating over all possible LLM pairs, we obtain a total of 14,400 action choices for the training datasets.

\subsection{Optimization}
\label{subsec:optimization}
In the phase of optimization, we optimize the parameters of the cognitive hierarchy models using the Maximum Likelihood Estimation (MLE). 
We formulate the problem as a constrained nonlinear optimization by applying the following log-likelihood function as our loss function: 
\begin{align*}
    \mathcal{L}= \sum_{(a_1^t,a_2^t)\in D_{G}} -\left[\log(s_1(a_1^t)) + \log(s_2(a_2^t))\right], 
\end{align*}
% \label{prog:LK}
where $D_{G}$ is the dataset of LLMs' action choices under game $G$, 
$a_1^t$ and $a_2^t$ are both agents' chosen actions in $t$-th round,
and $s_1(a_1^t)$ and $s_2(a_2^t)$ are the predicted overall strategies defined in \Cref{sec:prelim}.

We initialize the parameters of both the Level-$K$ CH and Poisson CH models randomly. The models' parameters are then estimated by minimizing the loss function $\mathcal{L}$ using Sequential Least Squares Programming \citep{kraft1988software}. Each optimization is repeated ten times, with the lowest-loss result reported as the final estimate.

\subsection{Evaluation}

With the optimized model parameters, for any given normal-form game with LLMs acting as the row player and column player respectively, we can employ cognitive hierarchy models to predict the potential level distributions of the two LLMs, as well as the specific strategy distributions at each level. These predicted results serve as the model's assessment and prediction of the LLMs' strategic reasoning capability.
Specifically, we care about these three questions:
\begin{itemize}
    \item \textbf{Q1:}
    In reality, people's strategic reasoning levels often have an upper bound. 
    How should the maximum level hyperparameter $\hat{k}$ be set? Is higher always better?
    \item \textbf{Q2:}
    Would an LLM exhibit consistent cognitive hierarchy levels when facing different opponents? 
    Does our framework demonstrate robustness and generalizability?
    \item \textbf{Q3:}
    What is the impact of the four distinct reasoning mechanisms on LLMs' strategic reasoning capability?
\end{itemize}

\begin{figure*}[!htb]
	\centering
    \begin{subfigure}{0.245\linewidth}
    \centering
    \includegraphics[width=\linewidth]{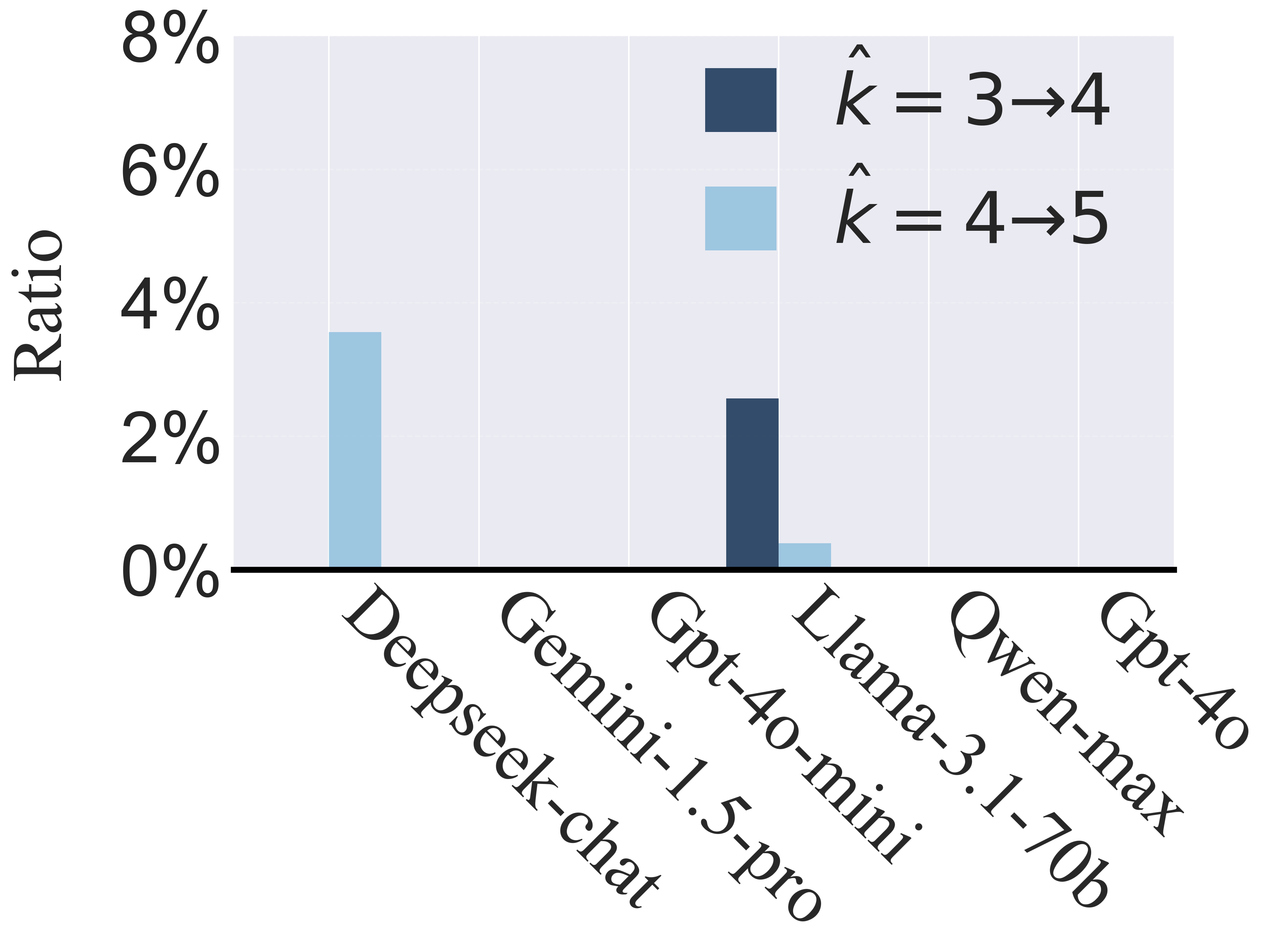}
    \caption{Baseline}
    \end{subfigure}
    	\begin{subfigure}{0.245\linewidth}
	\centering
	 \includegraphics[width=\linewidth]{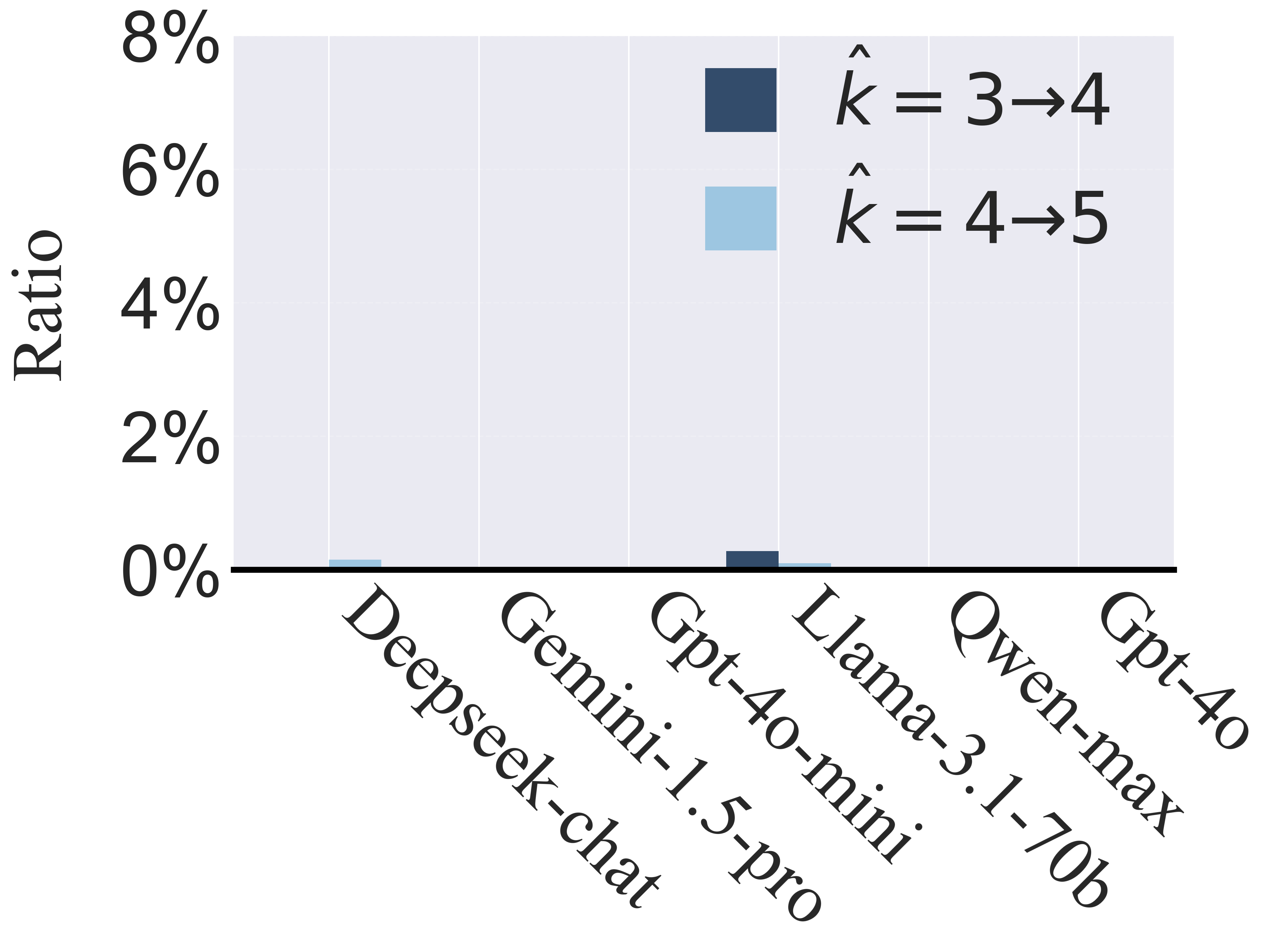}
	 \caption{Chat Mechanism}	
	\end{subfigure}
    \begin{subfigure}{0.245\linewidth}
    \centering
    \includegraphics[width=\linewidth]{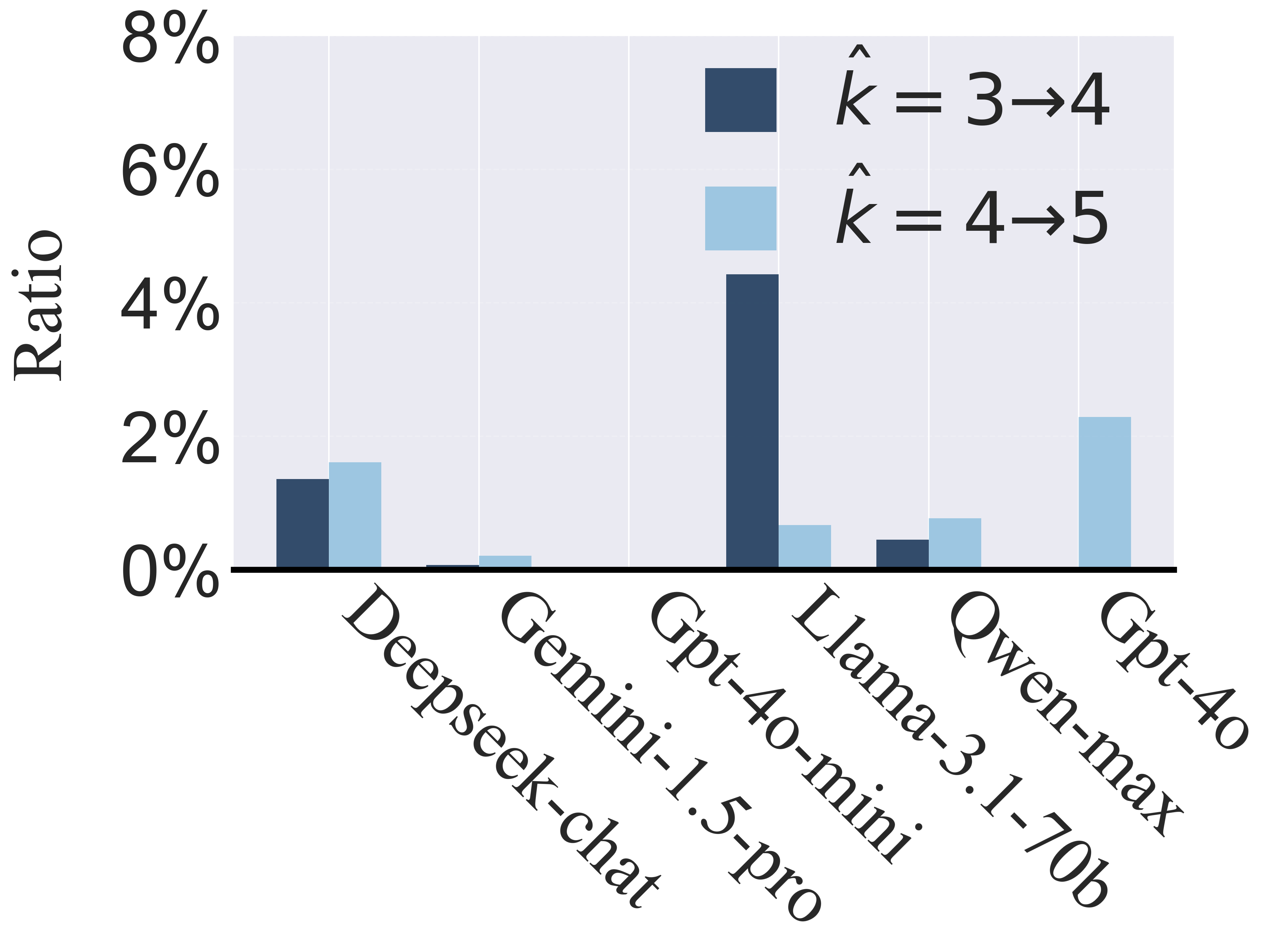}
    \caption{Memory Mechanism}
    \end{subfigure}
   	\begin{subfigure}{0.245\linewidth}
    \centering
    \includegraphics[width=\linewidth]{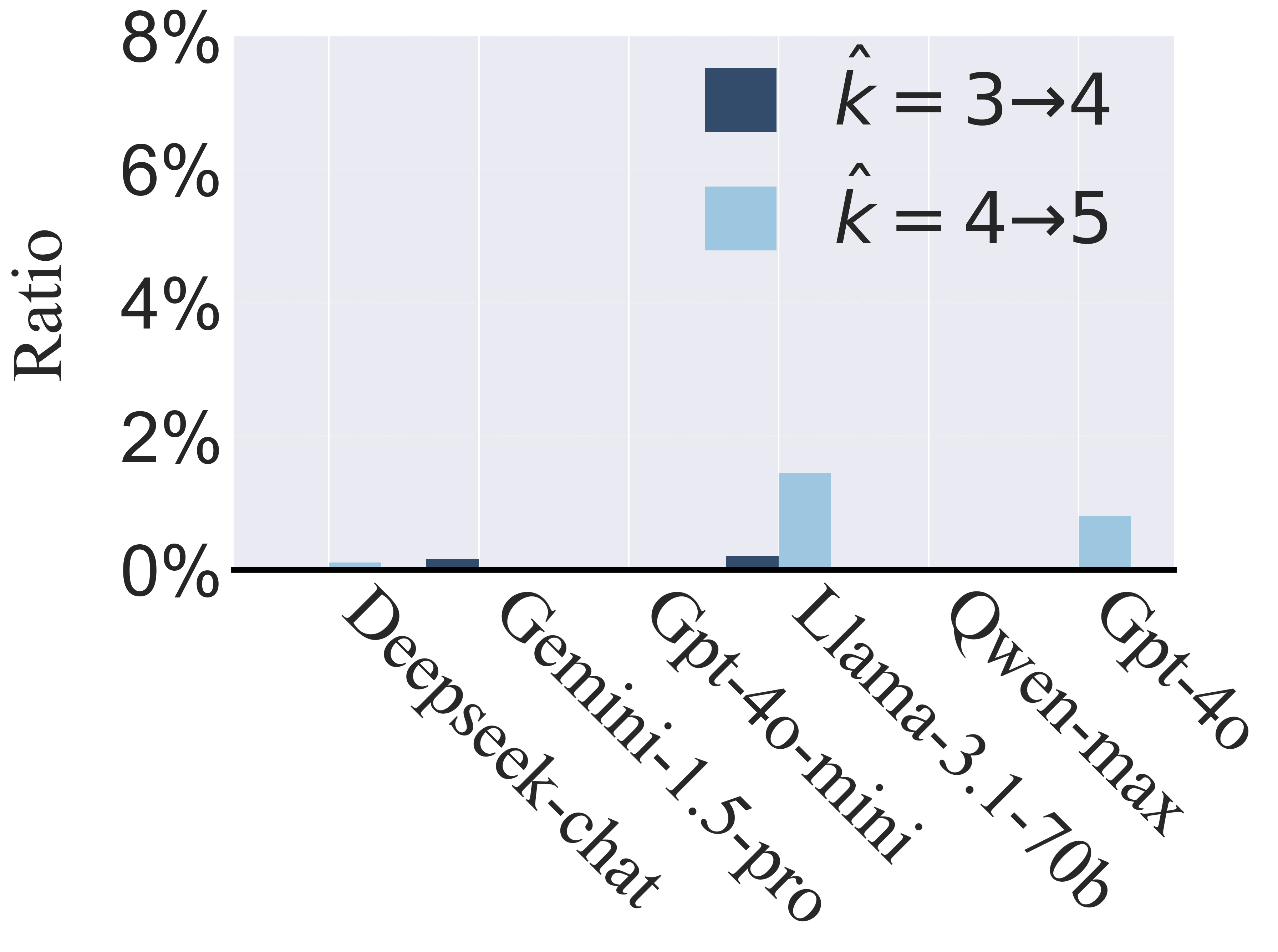}
    \caption{Chat \& Memory}
    \end{subfigure}
    \caption{Likelihood improvement ratio for maximum level $\hat{k}=3\rightarrow 4$ and $\hat{k}=4\rightarrow 5$ in the Level-K CH model.}
    \label{fig:exp_k_upper_bound_LK}
\end{figure*}
\begin{figure*}[!htb]
	\centering
    \begin{subfigure}{0.245\linewidth}
    \centering
    \includegraphics[width=\linewidth]{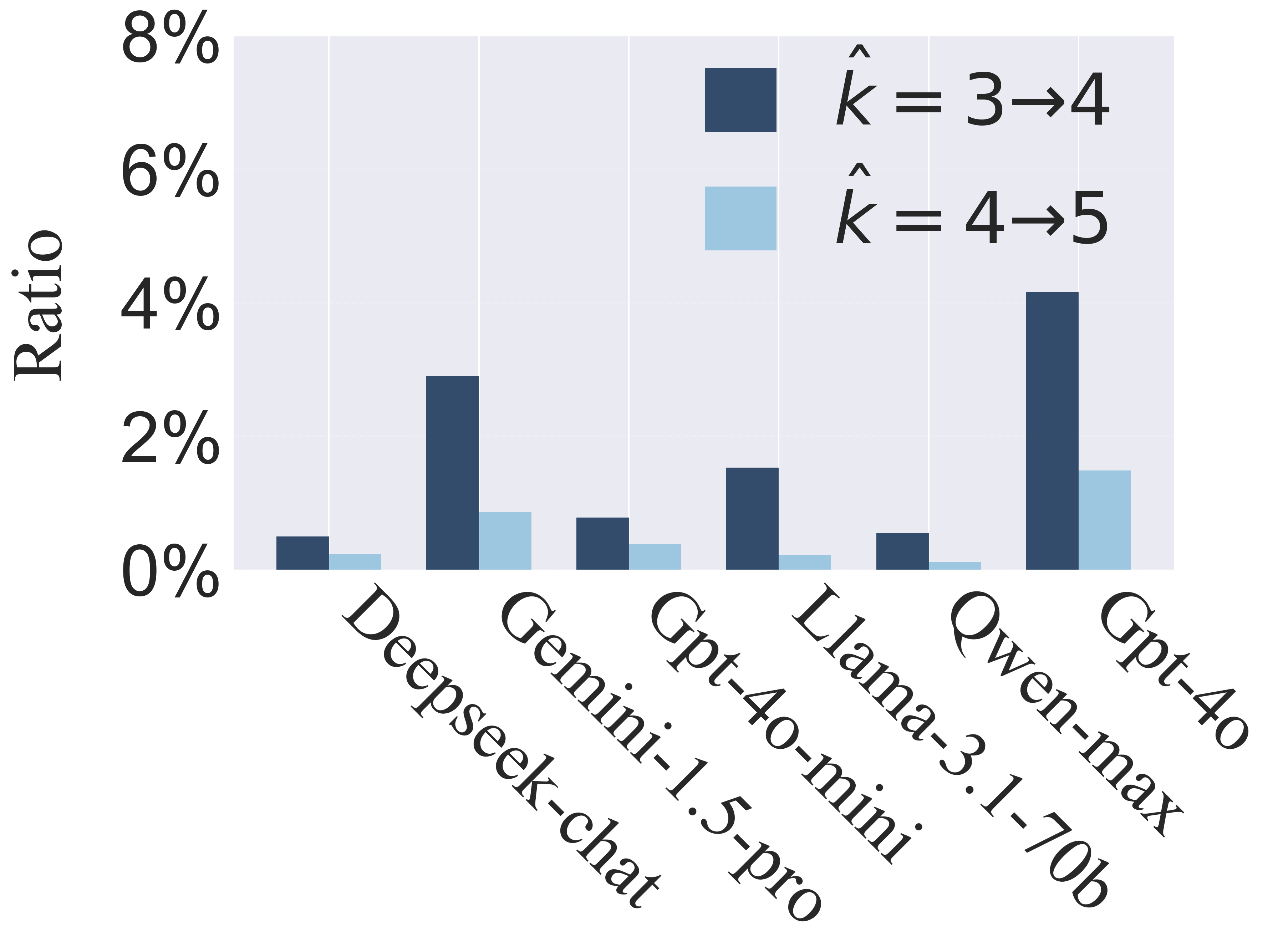}
    \caption{Baseline}
    \end{subfigure}
    	\begin{subfigure}{0.245\linewidth}
	\centering
	 \includegraphics[width=\linewidth]{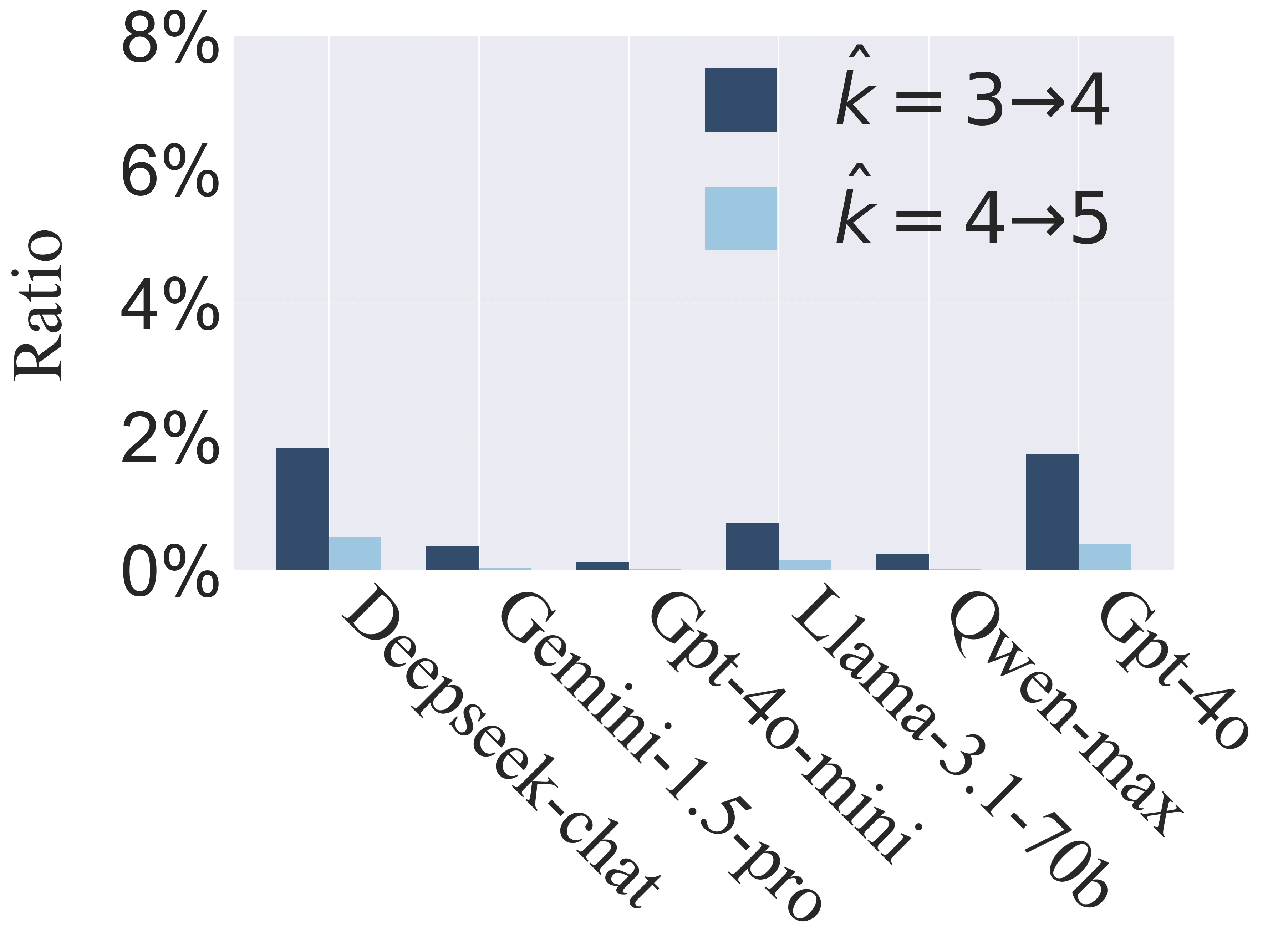}
	 \caption{Chat Mechanism}	
	\end{subfigure}
        \begin{subfigure}{0.245\linewidth}
    \centering
    \includegraphics[width=\linewidth]{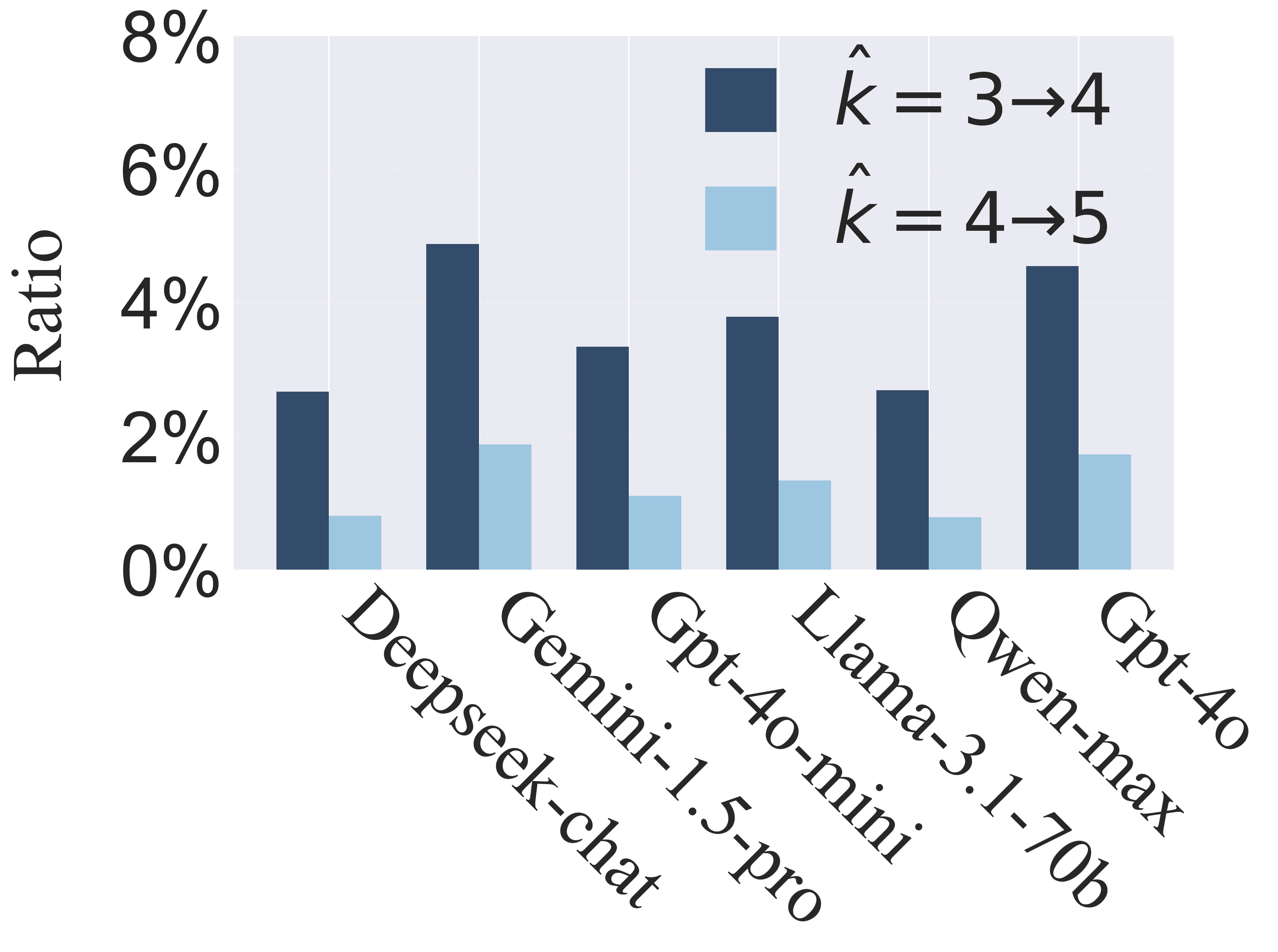}
    \caption{Memory Mechanism}
    \end{subfigure}
    	 \begin{subfigure}{0.245\linewidth}
    \centering
    \includegraphics[width=\linewidth]{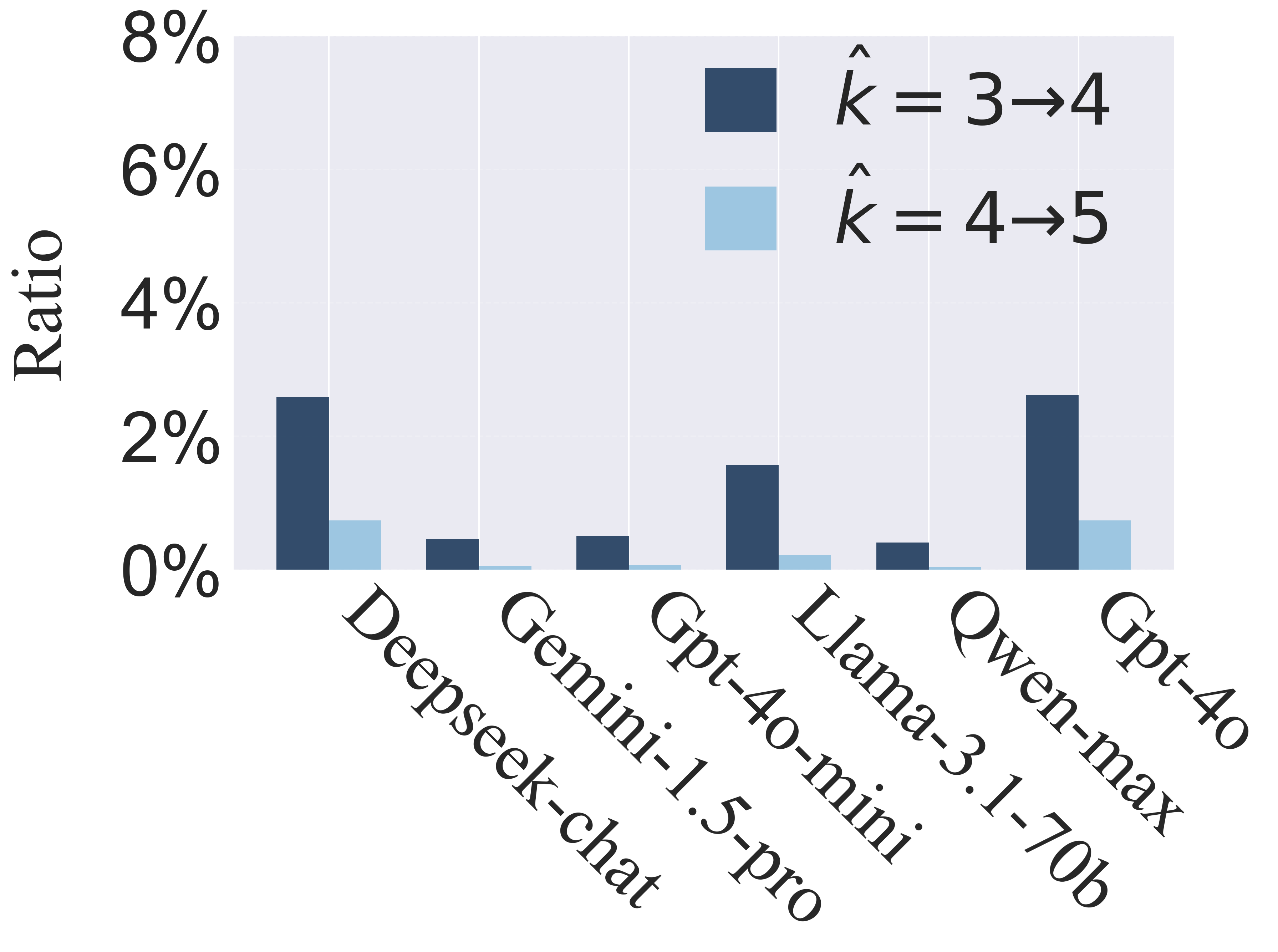}
    \caption{Chat \& Memory}
    \end{subfigure}
    \caption{Likelihood improvement ratio for maximum level $\hat{k}=3\rightarrow 4$ and $\hat{k}=4\rightarrow 5$ in the Poisson CH model.}
    \label{fig:exp_k_upper_bound_PCH}
\end{figure*}

\section{Experiment}
In this section, we aim to answer the above three questions. 
All experiments are run on a computer with an Intel Core i9-10940X CPU and 128 GB RAM.

\begin{table*}[htb!]
	\caption{Mean and variance of the predicted level distribution in the Level-K CH model under different reasoning mechanisms. Red cells indicate a negative effect and green cells indicate a positive effect. The bold cells represent the mechanisms with the best LLM strategy reasoning capability.}
\centering
	\resizebox{1\textwidth}{!}{
	\begin{tabular}{c|cc|cc|cc|cc}

	\multirow{2}{*}{\diagbox{\textbf{Model}}{\textbf{Mechanisms}}}		& \multicolumn{2}{c|}{\textbf{Baseline}}&\multicolumn{2}{c|}{\textbf{Chat Mechanism}}	&\multicolumn{2}{c|}{\textbf{Memory Mechanism}}&\multicolumn{2}{c}{\textbf{Chat \& Memory}}

	\\
		&\textbf{Var}	&\textbf{Mean}	&\textbf{Var}	&\textbf{Mean}	&\textbf{Var}   &\textbf{Mean}
		&\textbf{Var}  &\textbf{Mean}\\
		\Xhline{1pt}
DeepSeek-chat&0.0081&0.6639&0.0871&\cellcolor[RGB]{175, 240, 175}0.9490&0.0751&\cellcolor[RGB]{175, 240, 175}\textbf{1.4607}&0.0827&\cellcolor[RGB]{175, 240, 175}1.0765\\
Gemini-1.5-pro&0.0078&1.0530&0.0184&\cellcolor[RGB]{240, 175, 175}0.5713&0.0041&\cellcolor[RGB]{175, 240, 175}\textbf{1.0919}&0.0093&\cellcolor[RGB]{240, 175, 175}0.6140\\
GPT-4o-mini&0.0057&0.7381&0.0058&\cellcolor[RGB]{240, 175, 175}0.3016&0.0157&\cellcolor[RGB]{175, 240, 175}\textbf{1.0313}&0.0602&\cellcolor[RGB]{175, 240, 175}0.7608\\
Llama-3.1-70b&0.0074&1.1265&0.0314&\cellcolor[RGB]{240, 175, 175}0.5392&0.0394&\cellcolor[RGB]{175, 240, 175}\textbf{1.5417}&0.0144&\cellcolor[RGB]{240, 175, 175}0.9580\\
Qwen-max&0.0115&0.6761&0.0106&\cellcolor[RGB]{240, 175, 175}0.2707&0.0354&\cellcolor[RGB]{175, 240, 175}\textbf{1.1528}&0.0088&\cellcolor[RGB]{240, 175, 175}0.5641\\
GPT-4o&0.0087&1.1022&0.0055&\cellcolor[RGB]{240, 175, 175}0.6025&0.0097&\cellcolor[RGB]{175, 240, 175}\textbf{1.1754} &0.0204&\cellcolor[RGB]{240, 175, 175}0.9706\\
		\hline
	\end{tabular}
	}
	\label{tab:alpha_lk_var_mean_k4}
\end{table*}
\begin{table*}[htb!]
	\caption{Mean and variance of the predicted level distribution in the Poisson CH model under different reasoning mechanisms. Red cells indicate a negative effect and green cells indicate a positive effect. The bold cells represent the mechanisms with the best LLM strategy reasoning capability.}
\centering
	\resizebox{1\textwidth}{!}{
	\begin{tabular}{c|cc|cc|cc|cc}

	\multirow{2}{*}{\diagbox{\textbf{Model}}{\textbf{Mechanisms}}}	& \multicolumn{2}{c|}{\textbf{Baseline}}&\multicolumn{2}{c|}{\textbf{Chat Mechanism}}	&\multicolumn{2}{c|}{\textbf{Memory Mechanism}}&\multicolumn{2}{c}{\textbf{Chat \& Memory}}
	\\
		&\textbf{Var}	&\textbf{Mean}	&\textbf{Var}	&\textbf{Mean}	&\textbf{Var}   &\textbf{Mean}
		&\textbf{Var}  &\textbf{Mean}\\
		\Xhline{1pt}
DeepSeek-chat&0.0073&0.7139&0.0033&\cellcolor[RGB]{175, 240, 175}1.0506&0.0027&\cellcolor[RGB]{175, 240, 175}\textbf{1.1446}&0.0038&\cellcolor[RGB]{175, 240, 175}0.9832\\
Gemini-1.5-pro&0.0035&1.2152&0.0652&\cellcolor[RGB]{240, 175, 175}0.4603&0.0006&\cellcolor[RGB]{175, 240, 175}\textbf{1.3107}&0.0066&\cellcolor[RGB]{240, 175, 175}0.4157\\
GPT-4o-mini&0.0254&0.6061&0.0027&\cellcolor[RGB]{240, 175, 175}0.0513&0.0287&\cellcolor[RGB]{175, 240, 175}\textbf{1.1510}&0.0350&\cellcolor[RGB]{175, 240, 175}0.6935\\
Llama-3.1-70b&0.0002&0.7465&0.0074&\cellcolor[RGB]{240, 175, 175}0.6626&0.0006&\cellcolor[RGB]{175, 240, 175}\textbf{1.0651}&0.0253&\cellcolor[RGB]{240, 175, 175}0.6772\\
Qwen-max&0.0711&0.4362&0.0026&\cellcolor[RGB]{240, 175, 175}0.1963&0.0209&\cellcolor[RGB]{175, 240, 175}\textbf{1.2288}&0.0259&\cellcolor[RGB]{175, 240, 175}0.6294\\
GPT-4o&0.0012&1.2462&0.0041&\cellcolor[RGB]{240, 175, 175}0.6872&0.0022&\cellcolor[RGB]{175, 240, 175}\textbf{1.3192} &0.0415&\cellcolor[RGB]{240, 175, 175}1.2352   \\

		\hline
	\end{tabular}
	}
	\label{tab:p_ch_var_mean_k4}
\end{table*}

\subsection{Maximum Level $\hat{k}$}
\label{subsec:k_hat}
We argue that $\hat{k}=4$ adequately captures the strategic reasoning capability of LLMs. \citet{wright2010beyond, wright2017predicting} employ the ``likelihood improvement ratio'' to show that $\hat{k}=3$ effectively captures human strategic reasoning. We adopt the same metric to evaluate the $\hat{k}$ of LLMs.
Figures~\ref{fig:exp_k_upper_bound_LK} and~\ref{fig:exp_k_upper_bound_PCH} reveal substantial improvements from $\hat{k}=3\to4$, but minimal gains from $\hat{k}=4\to5$. In the Level-K CH model, most LLMs show no improvement from $\hat{k}=3\to4$, while in the Poisson CH model, they typically exhibit over 2\% improvement. 
Both models show marginal improvements beyond level-3. Thus, we set $\hat{k}=4$ across all experiments. whereas the $\hat{k}=5$ already exhibits overfitting.

Additionally, we also use KL-divergence to capture the difference of predicted strategies at different levels:
 \begin{align*}
     D_{KL}(s_i^{k},s_i^{k+1})=\sum_{a_i\in A_i}s_{i}^k(a_i)\log\frac{s_{i}^k(a_i)}{s_{i}^{k+1}(a_i)}.
 \end{align*}
The values of $D_{KL}(s_i^3,s_i^4)$ are almost zero across all LLMs, demonstrating that the distributions $s_i^3$ and $s_i^4$ in LLMs are nearly identical.
Due to the prevalence of zero values in $D_{KL}(s_i^3,s_i^4)$, all KL results are provided in the Supplementary Material.
Using both likelihood improvement ratios and KL-divergence methods, we argue that $\hat{k}=4$ sufficiently captures LLMs' strategic reasoning.
Thus, we set $\hat{k}=4$ across all experiments. whereas the $\hat{k}=5$ already exhibits overfitting.

\begin{table}
	\caption{Mean and variance of the predicted level distributions under partial memory mechanism.}
    \centering
	\begin{tabular}{c|cc|cc}
	\multirow{2}{*}{\textbf{Models}}	& \multicolumn{2}{c|}{\textbf{Level-K Model}}&\multicolumn{2}{c}{\textbf{Poisson CH Model}}\\
		&\textbf{Var}	&\textbf{Mean}	&\textbf{Var}	&\textbf{Mean}\\
		\Xhline{1pt}
DeepSeek-chat&0.0303&\cellcolor[RGB]{175, 240, 175}1.2060&0.0038&\cellcolor[RGB]{175, 240, 175}0.8393\\
Gemini-1.5-pro&0.0108&\cellcolor[RGB]{175, 240, 175}1.0014&0.0044&\cellcolor[RGB]{175, 240, 175}1.3619\\
GPT-4o-mini&0.0137&\cellcolor[RGB]{175, 240, 175}0.9920&0.0109&\cellcolor[RGB]{175, 240, 175}1.1457\\
Llama-3.1-70b&0.0207&\cellcolor[RGB]{175, 240, 175}1.4977&0.0022&\cellcolor[RGB]{175, 240, 175}0.9461\\
Qwen-max&0.0081&\cellcolor[RGB]{175, 240, 175}1.0733&0.0128&\cellcolor[RGB]{175, 240, 175}1.1735\\
GPT-4o&0.0083&\cellcolor[RGB]{175, 240, 175}1.2588&0.0146&\cellcolor[RGB]{175, 240, 175}1.3633 \\
		\hline
	\end{tabular}
	\label{tab:partial_memory_mechanism}
\end{table}

\subsection{Robustness of CHBench Framework} 

We use the variance of $k$ of LLM when facing different opponents to evaluate the robustness and generalization of the CHBench framework.
The LLM's strategic reasoning ability remains unchanged when facing unknown opponents and without additional information. As shown in \Cref{tab:alpha_lk_var_mean_k4} and \ref{tab:p_ch_var_mean_k4}, the variances in the $k$ values remain consistently below 0.1 in all experimental reasoning mechanisms within both Cognitive Hierarchy models. This low variance indicates that LLMs do not significantly alter their strategies when interacting with different players. So, we confirm that the CHBench framework is robust and generalized.

While the stability in reasoning patterns is evident, it's important to note that the average utility (payoff) of the LLMs displays considerable variance when facing different opponents. This suggests that despite maintaining consistent strategic approaches, the ultimate performance outcomes can still fluctuate depending on the specific interactions. %Due to space limitations, 
Detailed utility results are available in \Cref{sec:appendix_exp_lk} and \ref{sec:appendix_exp_u}.

\subsection{Strategic Reasoning Capability of LLMs}
\label{subsec:capabilities}
\Cref{tab:alpha_lk_var_mean_k4} and \ref{tab:p_ch_var_mean_k4} also show the strategic reasoning capabilities of LLM. In both cognitive hierarchy models, GPT-4o and Gemini-1.5-pro perform better when there is no external information. 

\Cref{tab:alpha_lk_var_mean_k4} and \ref{tab:p_ch_var_mean_k4} highlight the differential impact of mechanisms on the strategic reasoning capability of LLM.  
The Memory Mechanism consistently yields the most substantial enhancement in LLMs' strategic reasoning capability. This underscores the importance of historical game information in enabling LLMs to formulate more effective strategies.

A comparison between the baseline and the Memory Mechanism reveals divergent behavioral patterns across different LLMs. Models such as GPT-4o and Gemini-1.5-pro exhibit notable stability, with only marginal increments in their $k$ values (0.07 and 0.095, respectively) following the integration of historical information. This indicates that these models possess a robust dominant strategy that is less vulnerable to external perturbations. In contrast, LLMs with inferior inherent strategic reasoning capability demonstrate more pronounced strategic shifts and enhanced performance when furnished with historical context.

Conversely, the Chat Mechanism exerts a negative impact on LLMs' strategic reasoning capability, manifesting suboptimal performance. The Chat \& Memory Mechanism demonstrates marginal improvement over the Chat Mechanism alone, reinforcing that while access to historical information is beneficial, the Chat Mechanism tends to impair strategic reasoning. This detrimental effect may arise from an elevated propensity of cooperative models to adopt partners' suggestions, resulting in more cooperative yet potentially suboptimal strategies.

To further investigate the role of historical information, we conducted experiments with partial historical data (10 rounds) and the results compared to the full Memory Mechanism (30 rounds), as presented in \Cref{tab:partial_memory_mechanism}. The findings demonstrate a positive correlation between the amount of historical information and strategic reasoning performance. Moreover, even partial historical information outperforms the Chat \& Memory Mechanism, providing additional support for the importance of memory in strategic decision-making.

\begin{figure*}[!htb]
	\centering
	 \begin{subfigure}{0.245\linewidth}
    \centering
    \includegraphics[width=\textwidth]{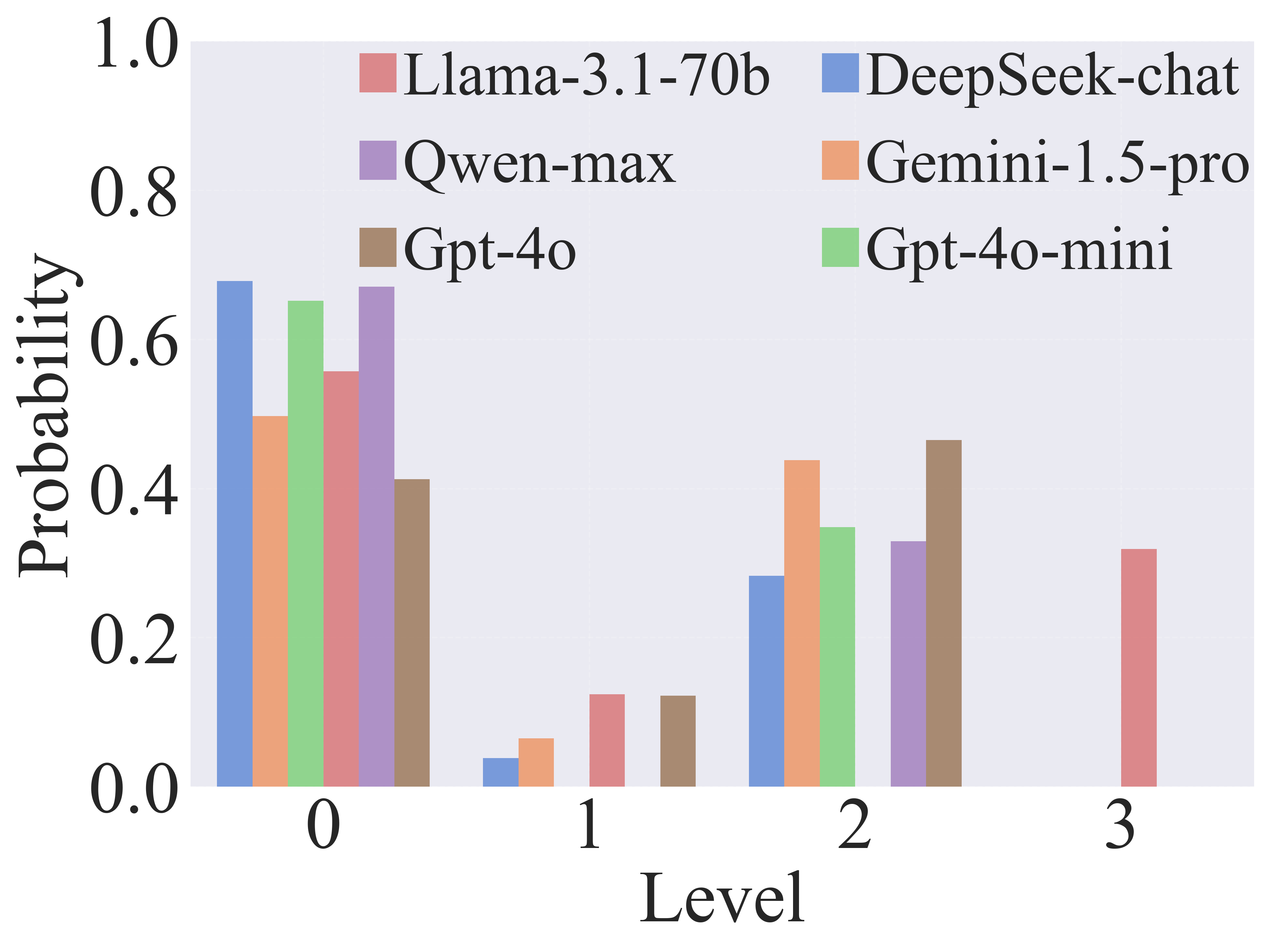}
    \caption{Baseline}
    \label{fig:exp_distribution_llm_a}
    \end{subfigure}
    \begin{subfigure}{0.245\linewidth}
	\centering
	 \includegraphics[width=\textwidth]{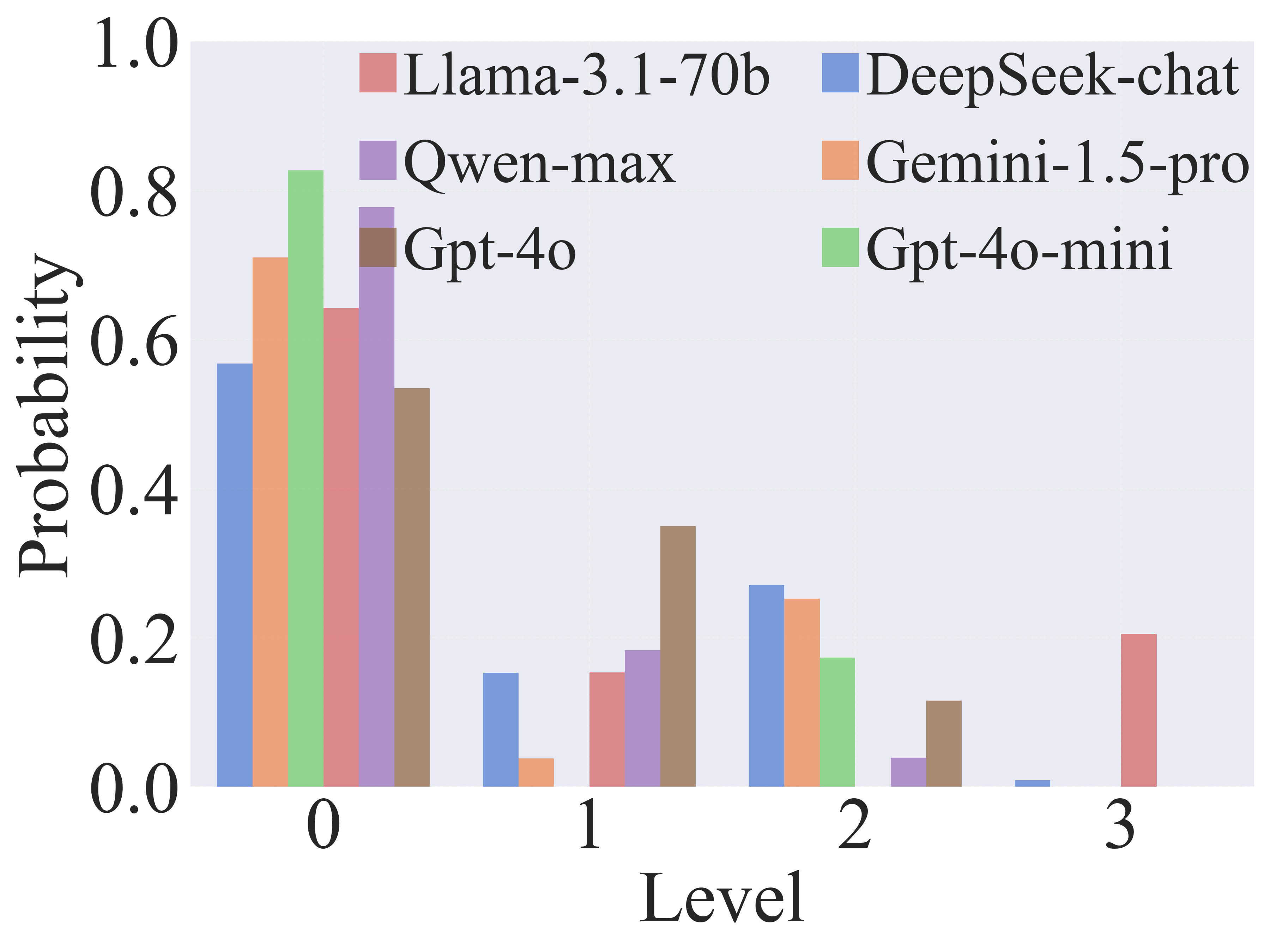}
	 \caption{Chat Mechanism}	
	 \label{fig:exp_distribution_llm_b}
	\end{subfigure}
    \begin{subfigure}{0.245\linewidth}
    \centering
    \includegraphics[width=\textwidth]{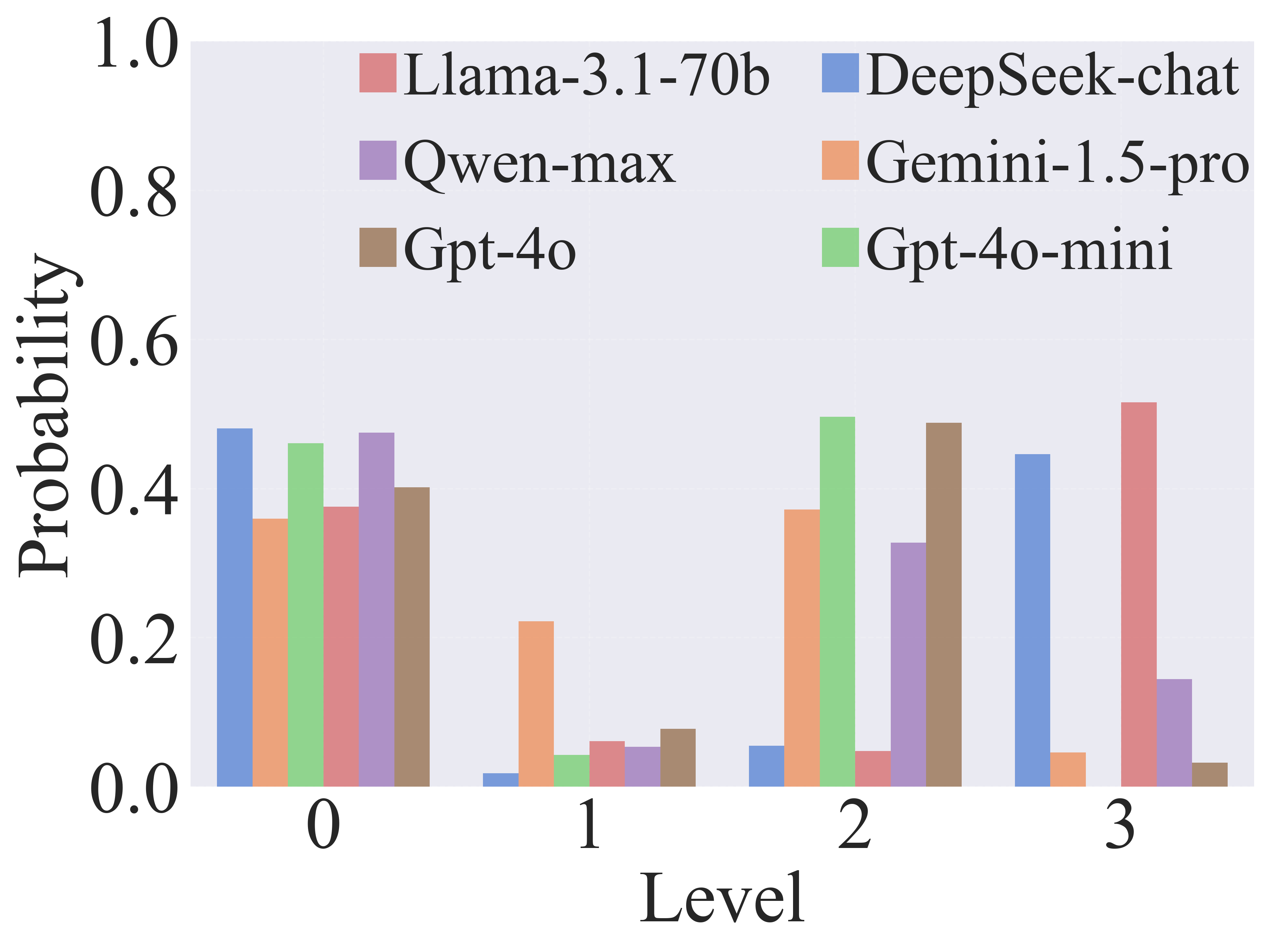}
    \caption{Memory Mechanism}
    \label{fig:exp_distribution_llm_c}
    \end{subfigure}
    \begin{subfigure}{0.245\linewidth}
    \centering
    \includegraphics[width=\textwidth]{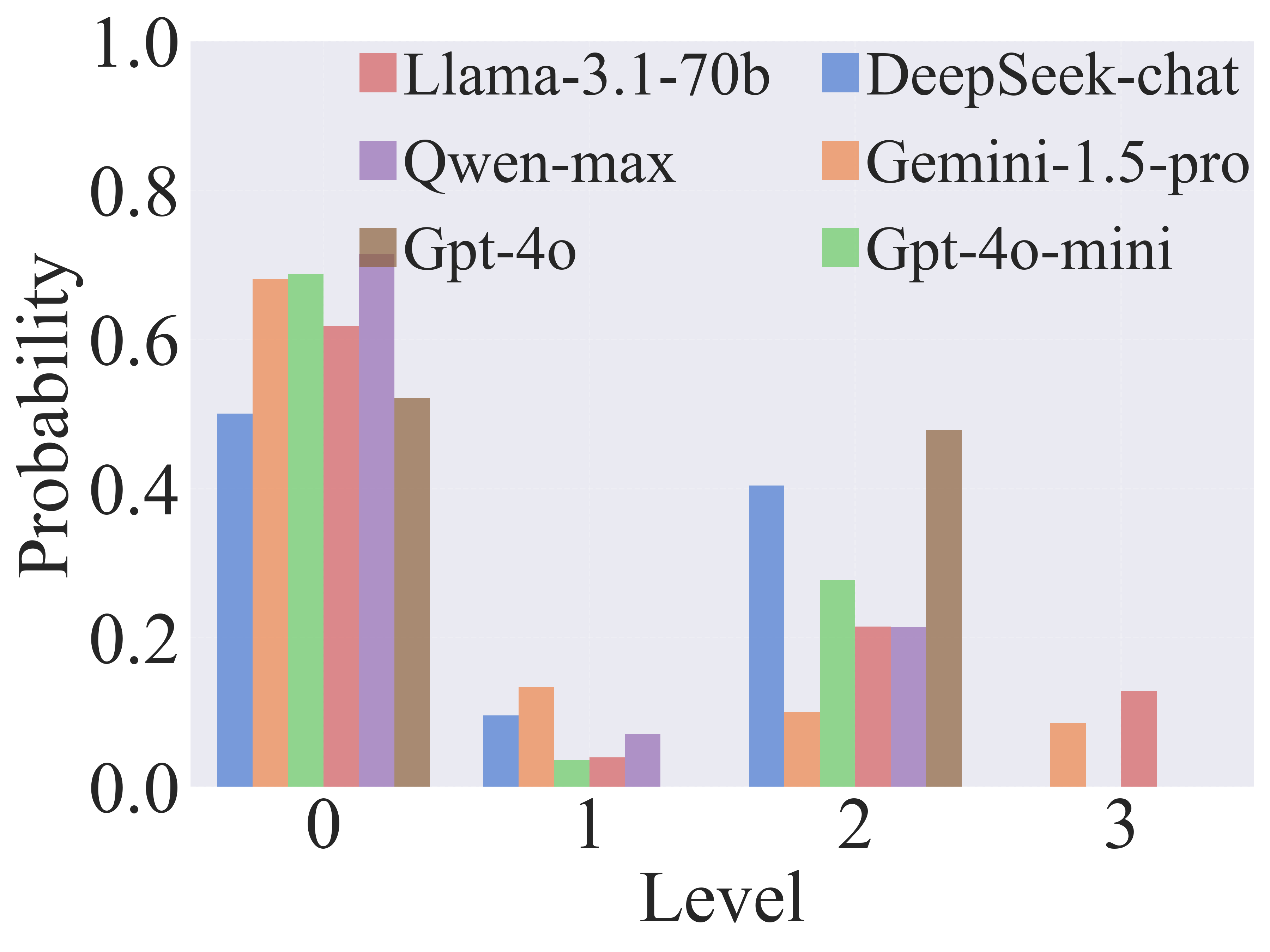}
    \caption{Chat \& Memory}
    \label{fig:exp_distribution_llm_d}
    \end{subfigure}
    \caption{Level distributions under different reasoning mechanisms in the Level-K CH model.}
    \label{fig:exp_distribution_llm_lk}
\end{figure*}
\begin{figure*}[!htb]
    \begin{subfigure}{0.245\linewidth}
    \centering
    \includegraphics[width=\textwidth]{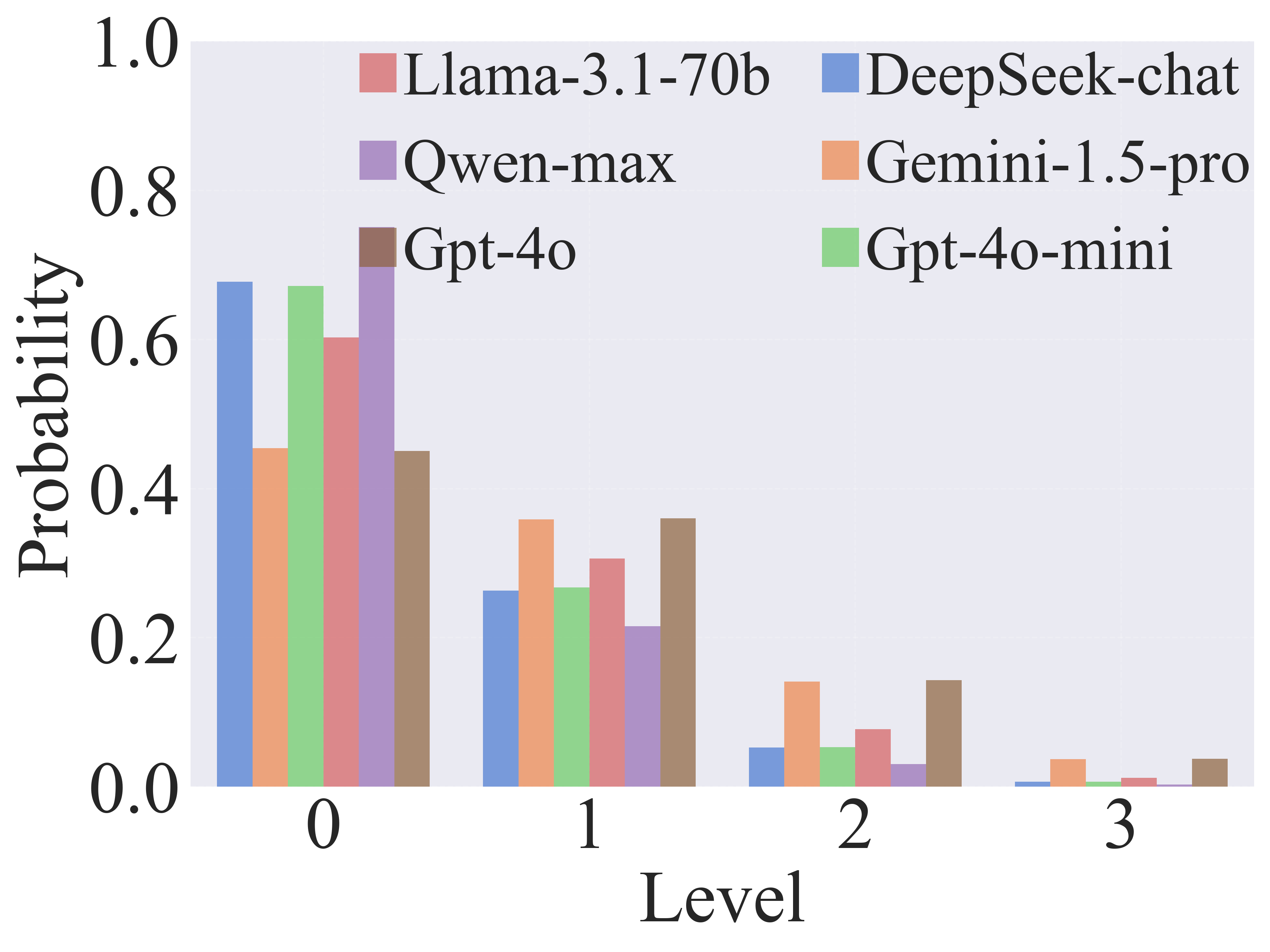}
    \caption{Baseline}
    \label{fig:exp_distribution_llm_e}
    \end{subfigure}
	\begin{subfigure}{0.245\linewidth}
	\centering
	 \includegraphics[width=\textwidth]{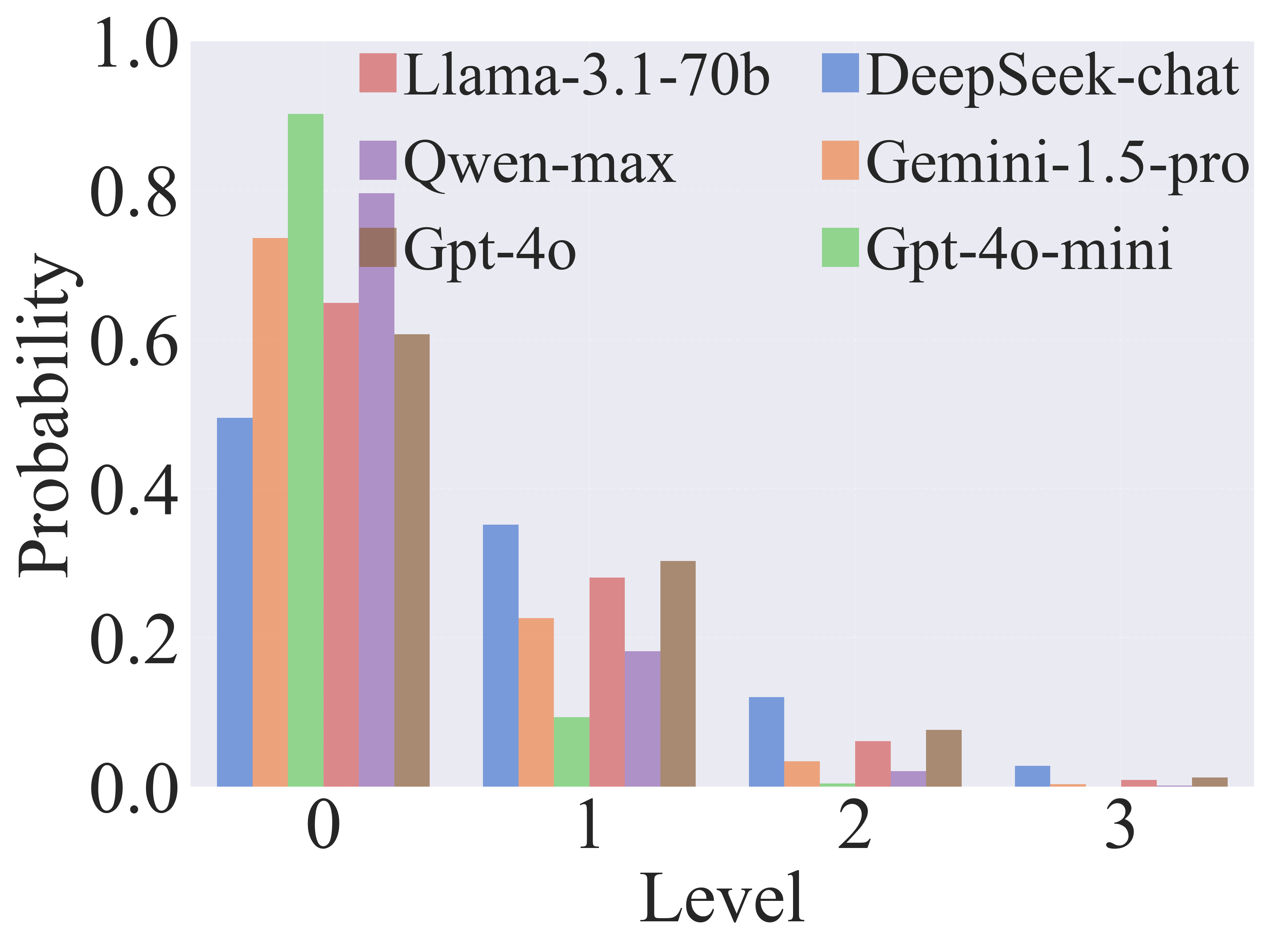}
	 \caption{Chat Mechanism}	
	 \label{fig:exp_distribution_llm_f}
	\end{subfigure}
    \begin{subfigure}{0.245\linewidth}
    \centering
    \includegraphics[width=\textwidth]{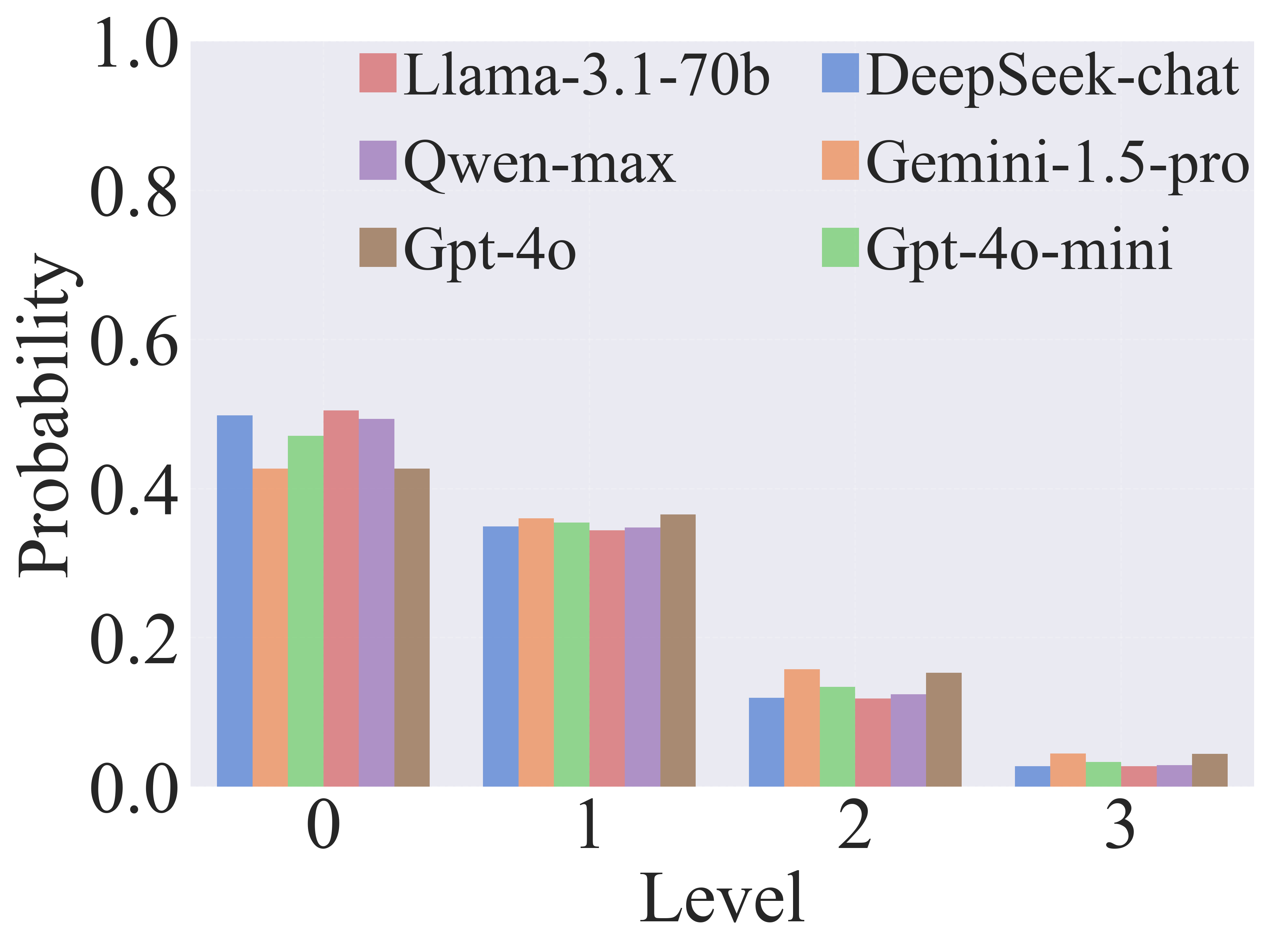}
    \caption{Memory Mechanism}
    \label{fig:exp_distribution_llm_g}
    \end{subfigure}
    \begin{subfigure}{0.245\linewidth}
	\centering
	 \includegraphics[width=\textwidth]{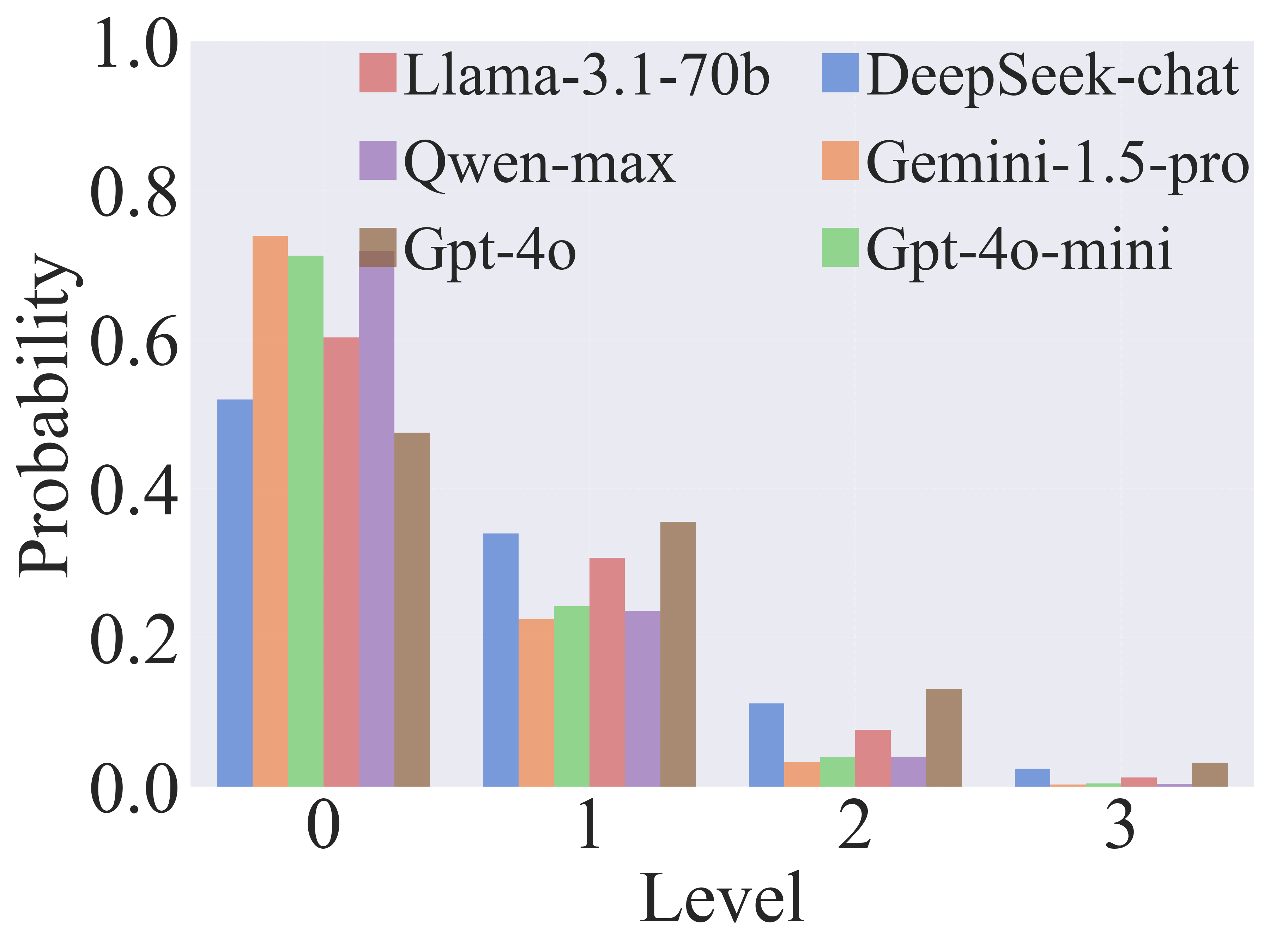}
	 \caption{Chat \& Memory}	
	 \label{fig:exp_distribution_llm_h}
	\end{subfigure}
    \caption{Level distributions under different reasoning mechanisms in the Poisson CH model.}
    \label{fig:exp_distribution_llm_pch}
\end{figure*}

\subsection{Predicted Level Distribution of LLMs}
\Cref{fig:exp_distribution_llm_lk} and \ref{fig:exp_distribution_llm_pch} illustrate the level distribution of LLMs, corroborating findings in \Cref{subsec:capabilities}. The results demonstrate that GPT-4o and Gemini-1.5-pro exhibit behavioral stability compared to other models, as evidenced by their minimal fluctuations in level distribution. 

LLMs could dynamically adjust their strategies based on historical information. Figures~\ref{fig:exp_distribution_llm_c} and~\ref{fig:exp_distribution_llm_g} show that the Memory Mechanism shifts level distributions, reducing the probability of $k=0$ while increasing the probability of $k>0$. 

In both cognitive hierarchy models, the Chat Mechanism assigns minimal probability to level-3. Specifically, the Level-K CH model concentrates at $k=0$ and $k=2$. The Memory Mechanism improves LLMs' strategic reasoning capabilities. LLMs show marked increases in $k>0$ probability. This confirms their capability enhanced for integrating historical information.

\label{sec:exp}

\section{Related Work}
\label{sec:related work}
\subsubsection*{Cognitive Hierarchy in Behavioral Economics}
Cognitive hierarchy models built upon normal-form games provide a framework for evaluating human decision-making under bounded rationality. 
Previous research has extensively applied these models to study human strategic behavior. For example, \citet{stahl1995players} examined human performance in symmetric games $3\times3$, while \citet{costa2001cognition} proposed a classification of five types of strategies to assess human strategic sophistication. 
Empirical studies of human gameplay data, using the Level-K CH and Poisson CH models, have advanced understanding of strategic reasoning.\citep{camerer2004cognitive,chong2016generalized,wright2010beyond,wright2014level,wright2017predicting,wright2020formal}. 
%Subsequent research \citep{traulsen2010human,camerer2015cognitive,wright2020formal} contrasts strategic and non-strategic decision-making, uncovering cognitive drivers of human behavior.

\subsubsection*{Game-theoretical Evaluation of LLMs}
Recent studies have systematically evaluated LLMs as game-theoretic agents through progressively sophisticated approaches. The initial work of \citet{akata2023playing} established a foundational benchmark using normal-form games. %, while the subsequent research by \citet{liga2023testing,topsakal2024benchmarking,vonderlind2024automatically} employed simplified environments such as Tic-Tac-Toe to measure basic strategic reasoning capabilities. 
Researchers have further examined LLM behavior through canonical games that reveal specific strategic tendencies: \citet{fontana2023llm,fontana2024nicer} compared LLM and human strategies in the Prisoner's Dilemma. %, while \citet{lore2023strategic,hua2024game} analyzed cooperative behaviors via Stag Hunt scenarios. 
However, these studies consistently highlight key limitations in the strategic reasoning of LLMs, including suboptimal performance and vulnerability to deception \citep{herr2024large,taylor2025large}. To address these challenges, recent work has developed diverse benchmarking frameworks - from the 10-game suite by \citet{duan2024gtbench} %and amic scenarios by \citet{huang2024far} to language-based sequential games by \citet{shapira2024glee} 
and topology-based evaluations by \citet{wang2024tmgbench}. Comprehensive reviews by \citet{sun2025game} and \citet{zhang2024llm} synthesize these methodological advances, providing structured frameworks for future research in LLM strategic reasoning.

\section{Conclusion}
\label{sec:conclusion}

% Recent advances in LLMs have demonstrated strong decision-making and reasoning capabilities. 
% However, their inherent instability poses challenges in identifying appropriate metrics and evaluation methods. 
%Inspired by the Cognitive Hierarchy model from behavioral economics, 
We propose a CHBench framework for evaluating the strategic reasoning capability of LLMs inspired by the cognitive hierarchy models. 
%Leaving behind the assumption of perfect rationality in classical game theory, our approach accounts for bounded rationality, where agents operate at different depths of strategic reasoning. 
This framework effectively captures the decision-making instability observed in LLMs.
Experimental results show that the CHBench framework robustly evaluates strategic reasoning capability while remaining resilient to variations in opponent strategies. 
%As a preliminary application, we used the CHBench framework to assess two distinct mechanisms, revealing their differential impacts on reasoning performance. 
We believe that this framework establishes a new paradigm for LLM evaluation. 

Two interesting future works can be studied: (1) investigating the integration of the CHBench framework into the Chain-of-Thought (CoT) reasoning paradigm, and (2) evaluating whether CoT’s strategic reasoning capability is higher when it predicts opponents' actions more accurately during its thinking process. Key challenges include ensuring seamless compatibility between CHBench framework and CoT’s reasoning mechanisms.

\bibliographystyle{abbrvnat}
\bibliography{arxiv_ref.bib}

\clearpage
\appendix
\section{Appendix}
\subsection{Prompt Templates}
\label{sec:prompt_templates}
In this section, we present the prompt templates we used for constructing the LLM-based agents and the game-playing environment. 
Across different reasoning mechanisms, the descriptions of game rules including action sets and possible outcomes remain consistent in prompts. 
For each mechanism, new descriptions are introduced to the baseline prompt to form a modified version. 
\begin{figure}[!htb]
	\centering
    \begin{subfigure}{\linewidth}
    \centering
    \includegraphics[width=\linewidth]{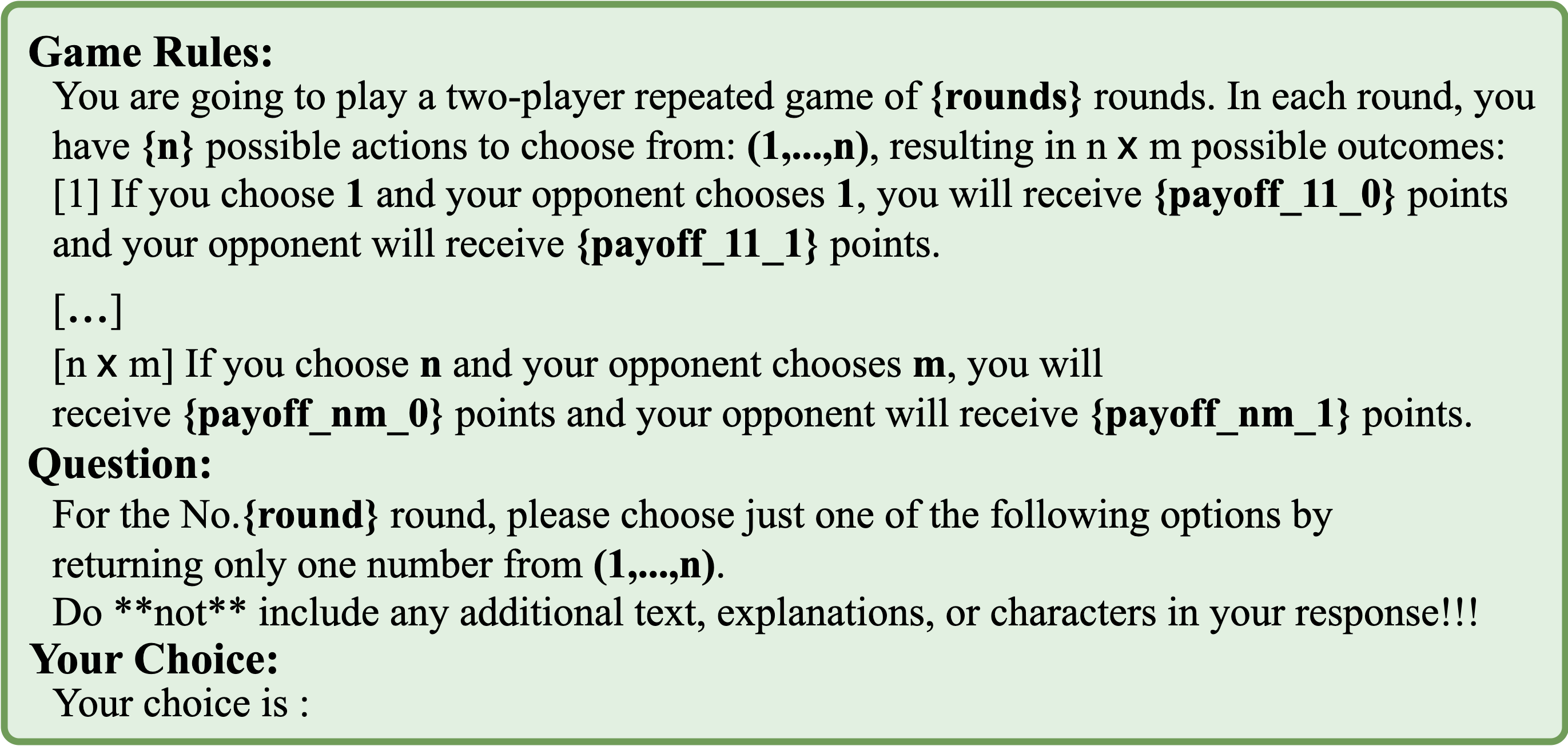}
    \end{subfigure}
    \caption{Decision prompt template under the Baseline Mechanism.}
    \label{fig:prompt_no_memory}
\end{figure}
\begin{figure}[!htb]
	\centering
    \begin{subfigure}{1\linewidth}
    \centering
    \includegraphics[width=\linewidth]{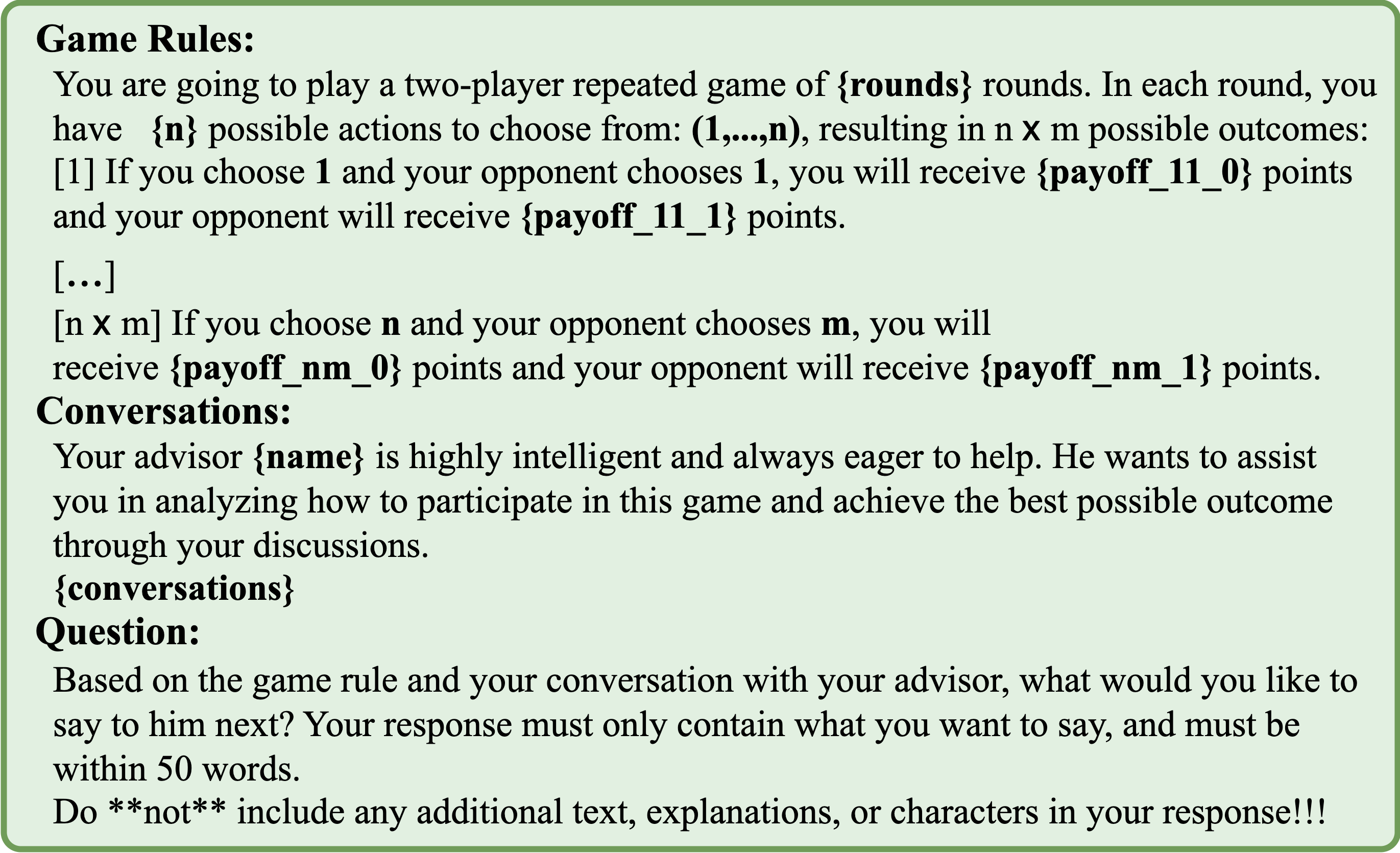}
    \end{subfigure}
    \caption{Consultation prompt template under the Chat Mechanism.}
    \label{fig:prompt_chat_conversation}
\end{figure}
\begin{figure}[!htb]
	\centering
    \begin{subfigure}{1\linewidth}
    \centering
    \includegraphics[width=\linewidth]{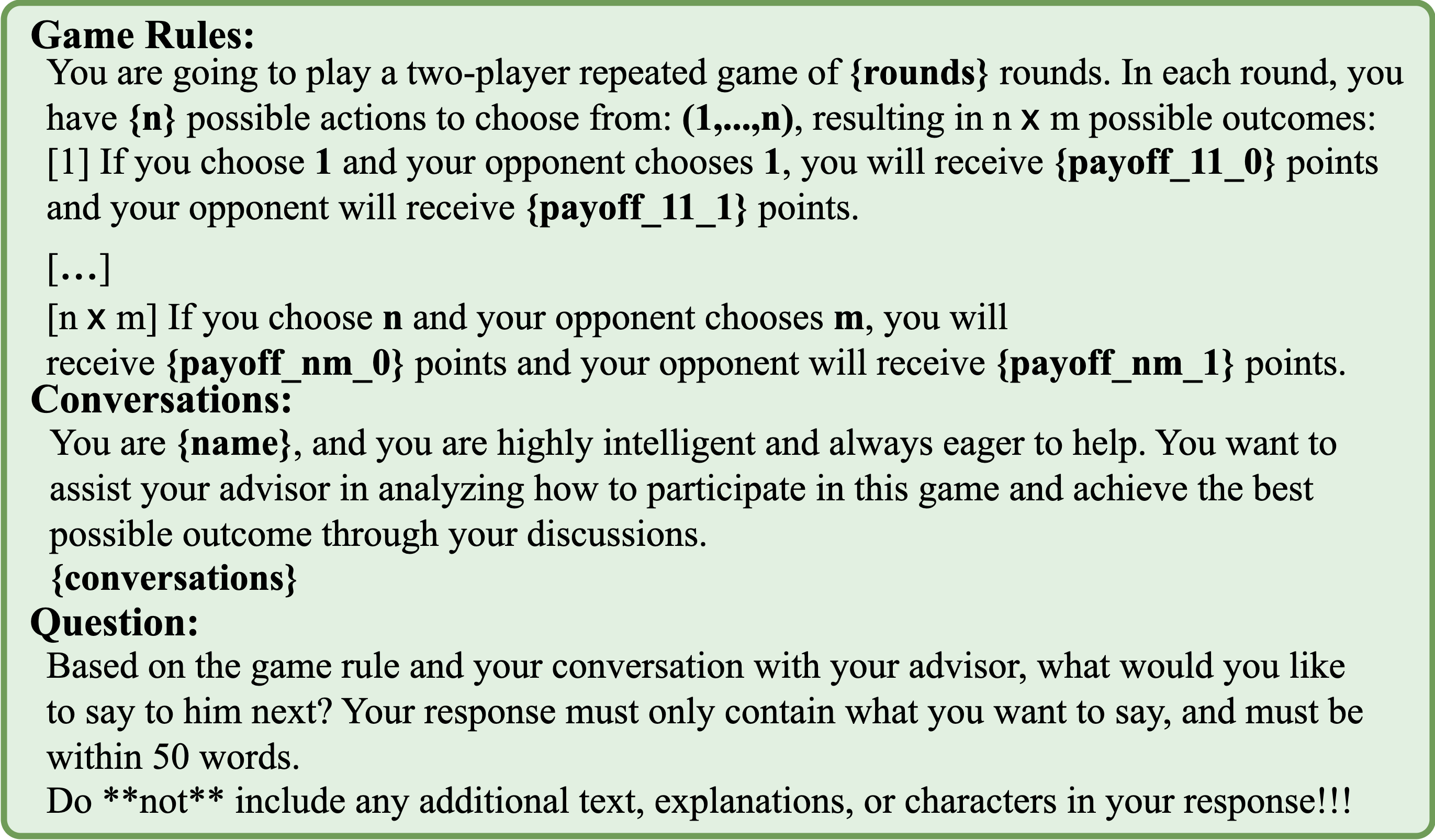}
    \end{subfigure}
    \caption{Response prompt template under the Chat Mechanism.}
    \label{fig:prompt_chat_friend_response}
\end{figure}
\begin{figure}[!htb]
         \begin{subfigure}{1\linewidth}
    \centering
    \includegraphics[width=\linewidth]{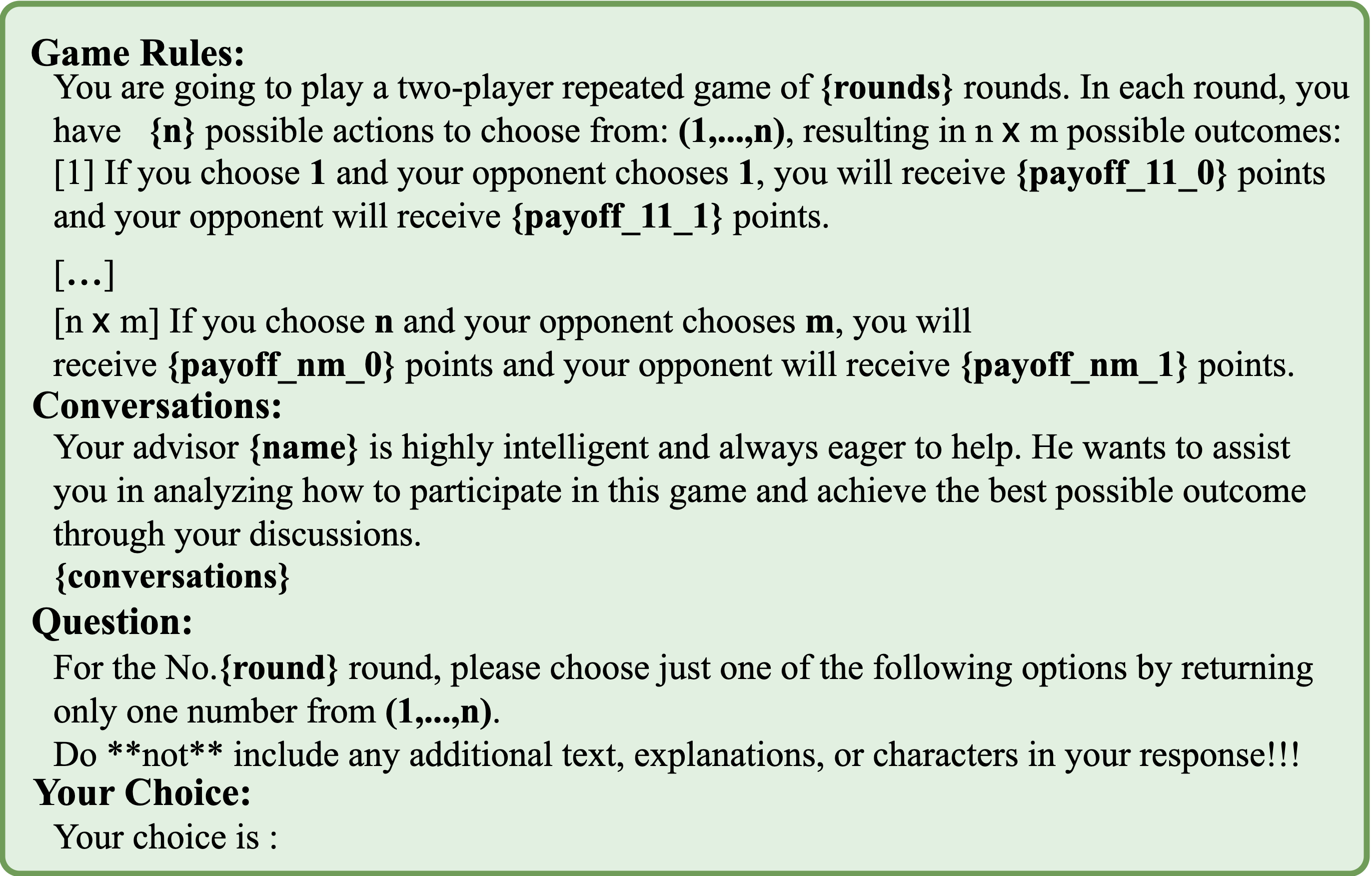}
    \end{subfigure}
        \caption{Decision prompt template under the Chat Mechanism}
    \label{fig:prompt_chat_mechanism_choose}
\end{figure}
\begin{figure}[!htb]
	\centering
    \begin{subfigure}{1\linewidth}
    \centering
    \includegraphics[width=\linewidth]{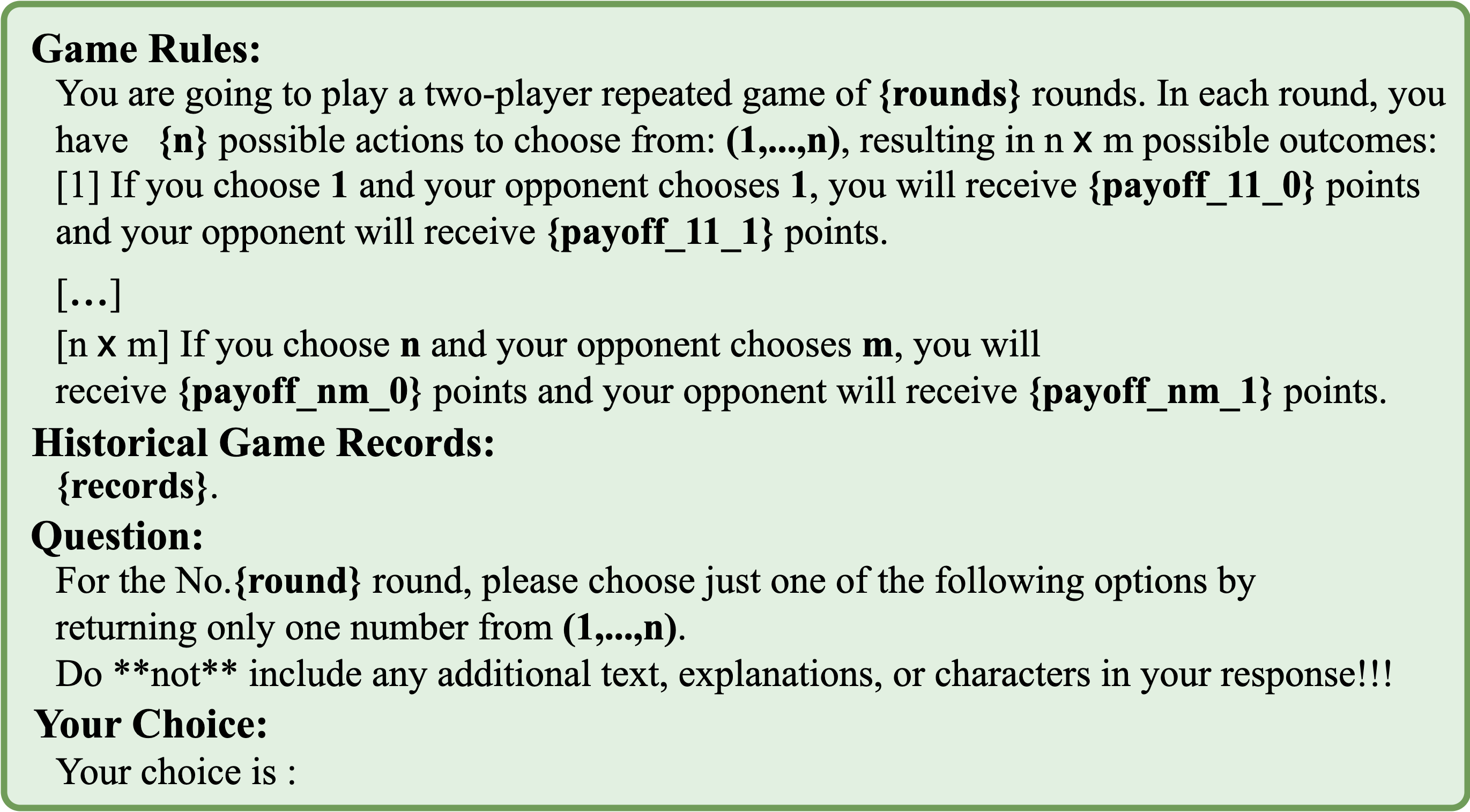}
    \end{subfigure}
    \caption{Decision prompt template under the Memory Mechanism.}
    \label{fig:prompt_memory_mechanism}
\end{figure}
\begin{figure}[!htb]
	\centering
    \begin{subfigure}{1\linewidth}
    \centering
    \includegraphics[width=\linewidth]{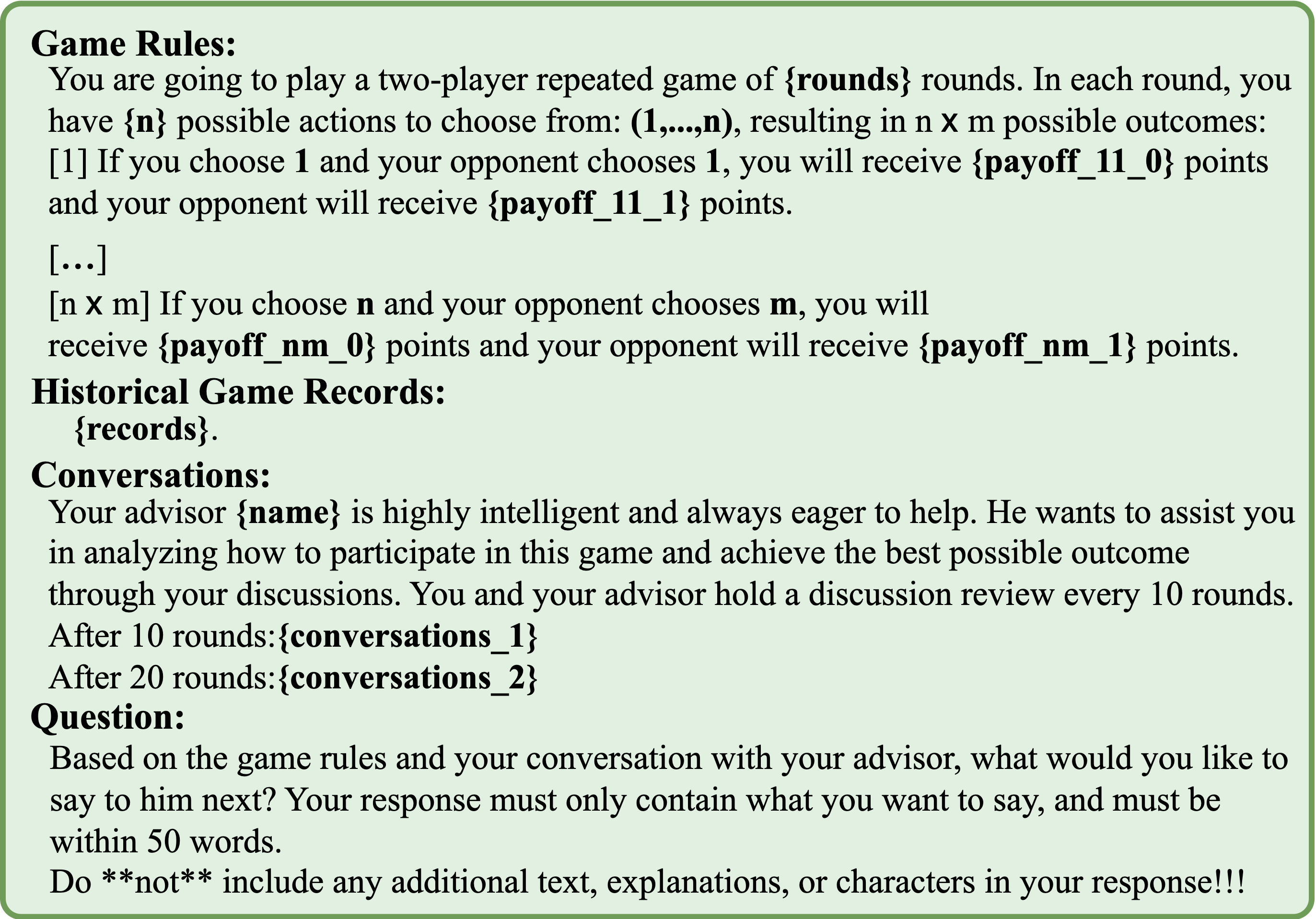}
    \end{subfigure}
    \caption{Consultation prompt template under the Chat \& Memory Mechanism.}
    \label{fig:prompt_chat_memory_mechanism}
\end{figure}
\begin{figure}[!htb]
	\centering
    \begin{subfigure}{1\linewidth}
    \centering
    \includegraphics[width=\linewidth]{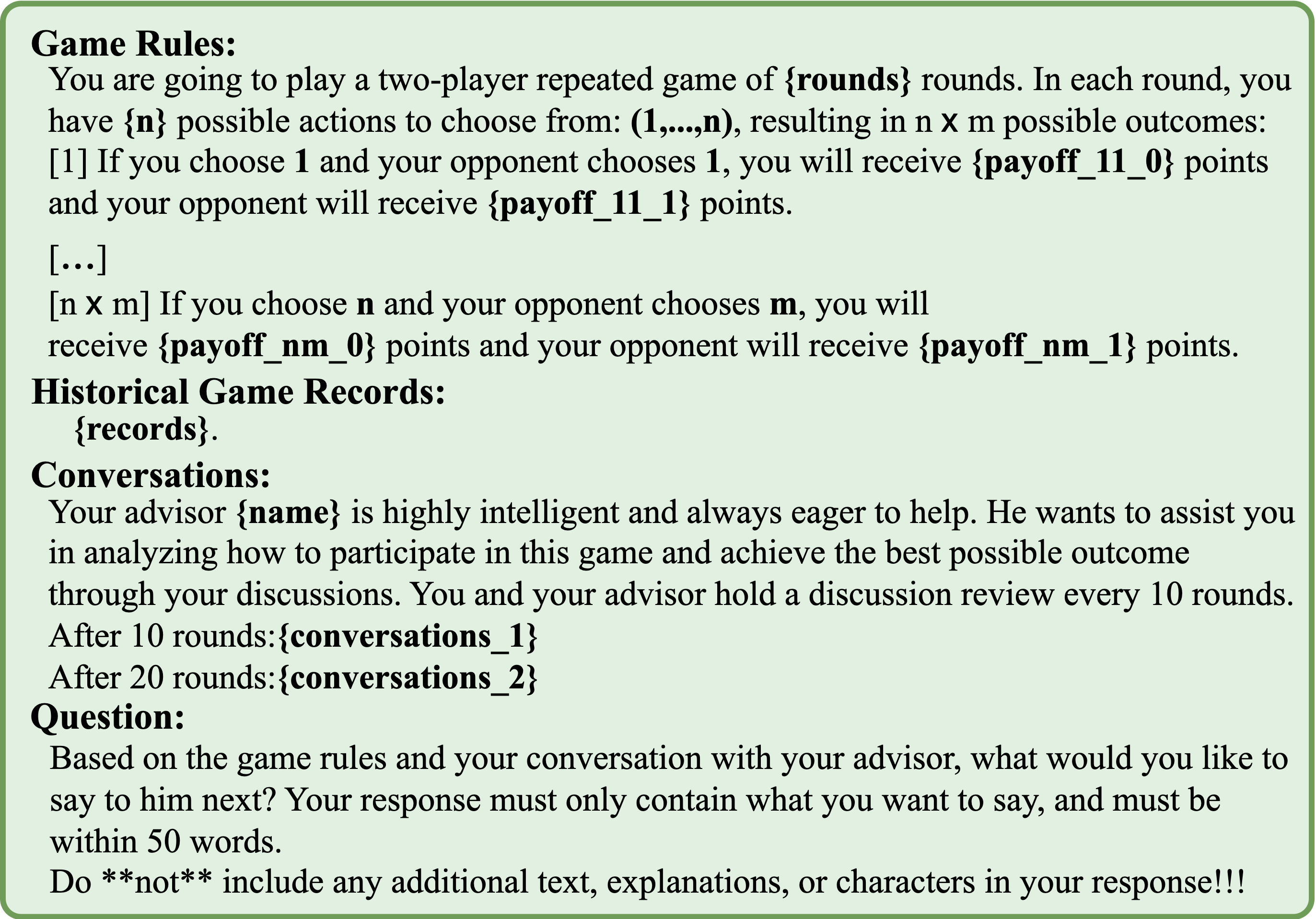}
    \end{subfigure}
    \caption{Response prompt template under the Chat \& Memory Mechanism.}
    \label{fig:prompt_chat_memory_friend_conversation}
\end{figure}
\begin{figure}[!htb]
         \begin{subfigure}{1\linewidth}
    \centering
    \includegraphics[width=\linewidth]{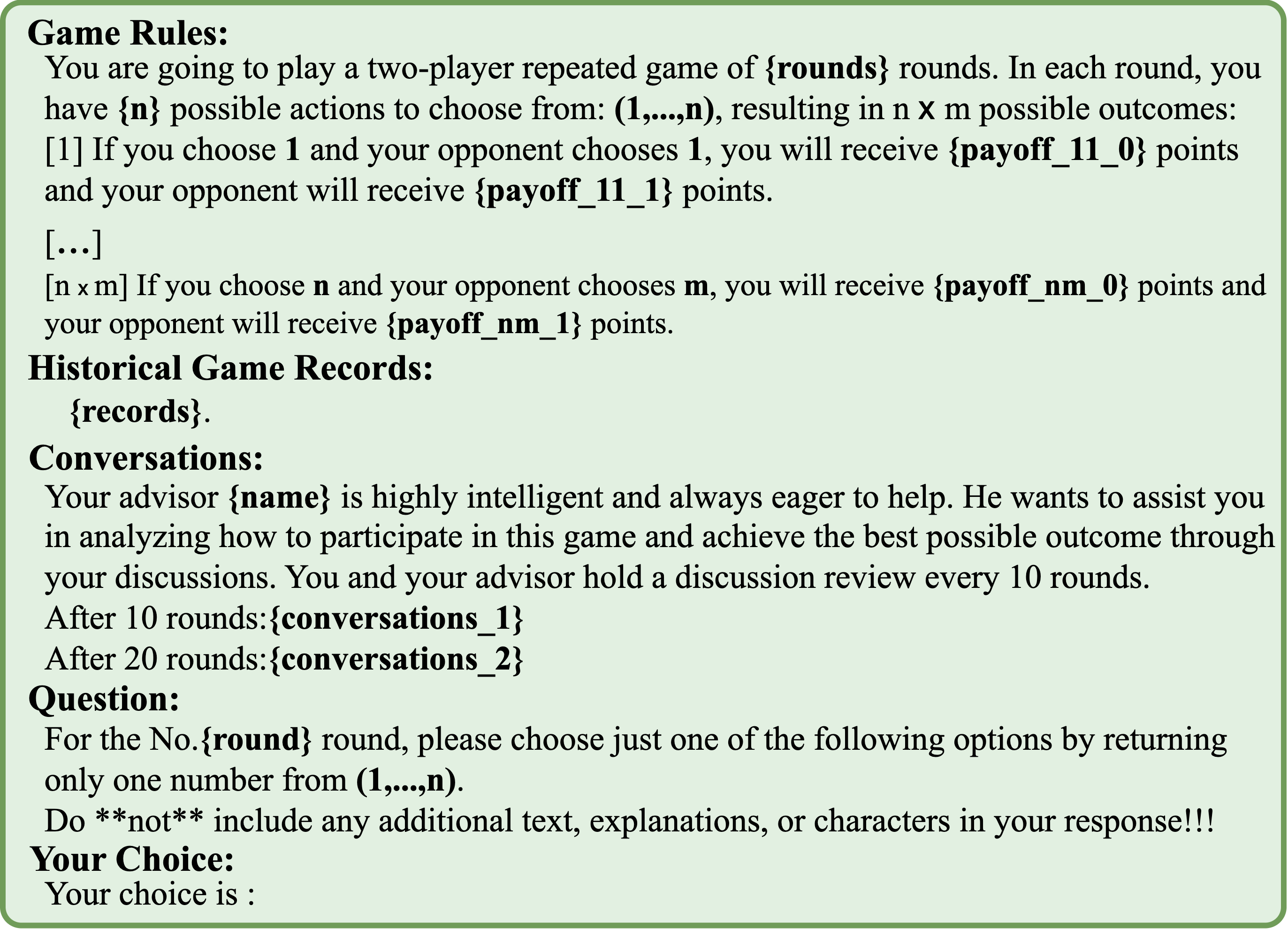}
%    \caption{{ }
    \end{subfigure}
        \caption{Decision prompt template under the Chat \& Memory Mechanism}
    \label{fig:prompt_chat_memory_choose}
\end{figure}

In the following prompt templates, the content inside the curly braces represents parameters that need to be replaced based on actual circumstances.
In the \textbf{Game Rules} and \textbf{Question }section shared by all templates, parameter \textbf{\{n\}} represents the number of actions available to the agent, and \textbf{\{m\}} represents the number of actions available to the agent's opponent. 
Depending on the number of actions available to both players, there are $n\times m$ possible outcomes in total. 
Each outcome clearly specifies the actions chosen by both parties and the resulting payoffs.
\textbf{\{payoff\_ij\_0\}} denotes the payoff for the agent when he chooses the $i$-th action and its opponent selects the $j$-th action, while \textbf{\{payoff\_ij\_1\}} represents the opponent's payoff under the same circumstances. 
Parameter \textbf{\{rounds\}} specifies the number of game repetitions in total, and parameter \textbf{\{round\}} refers to the current round number.

\subsubsection*{Baseline Mechanism}
Under the Baseline Mechanism, there is only one kind of prompt template, that is the Decision prompt template shown in \Cref{fig:prompt_no_memory} below.

Our baseline prompt is structured into three key components: (1) \textbf{Game Rules}, which outline the gameplay mechanics and specify the payoff structure for all possible outcomes given the players' actions; (2) \textbf{Question}, which presents the task the LLM needs to answer along with the required response format; and (3) \textbf{Your Choice}, where the LLM provides its final decision.
Based on this prompt template, the LLM needs to comprehend the game rules and internal game-theoretic relationships through textual understanding, then select an action from the action space $\{1,\dots,n\}$ and return the action chosen.

\subsubsection*{Chat Mechanism}
Under the Chat Mechanism, there is three kinds of prompt templates, that is the Consultation prompt template of the agent (\Cref{fig:prompt_chat_conversation}), the Response prompt template of his advisor (\Cref{fig:prompt_chat_friend_response}), and the Decision prompt template of the agent (\Cref{fig:prompt_chat_mechanism_choose}).

By using the Consultation prompt template, the agent seeks advice from his advisor regarding both the game rules and decision-making strategies. 
This template incorporates an additional \textbf{Conversations} module that stores the complete chat history between the agent and its advisor. 
The interaction follows a structured two-round communication protocol, where each question-and-answer exchange constitutes one full round of consultation.

We characterize the advisor as an intelligent and helpful individual who provides advice in favor of the agent's interests. The advisor communicates with the agent using the aforementioned Response prompt template, while the \textbf{Conversations} module stores the complete chat history between the agent and its advisor.
Notably, we assign distinct names to both the agent and its advisor. This design serves two crucial purposes: (1) enabling the LLM to clearly understand whose perspective it should adopt during reasoning and decision-making, and (2) preventing pronoun confusion (e.g., ambiguity about whether "you" refers to the agent itself or its counterpart). Without such naming conventions, the LLM would frequently misinterpret personal references during the interaction.

After completing two rounds of conversation between each agent and its respective advisor, each agent must then make specific action choices for each round using the aforementioned Decision prompt template. When making decisions, the agent needs to incorporate information from the \textbf{Conversations} module, which may include discussions about game rules, strategic planning, and other relevant content. The agent must carefully weigh and choose between its own reasoning and the dialogue content before making its final decision.

\subsubsection*{Memory Mechanism}
In the Memory Mechanism, there is only one kind of prompt template, that is the decision prompt template shown in \Cref{fig:prompt_memory_mechanism} below.

Building upon the decision prompt template under the Baseline Mechanism (see \Cref{fig:prompt_no_memory}), we incorporate additional components to enhance the model's reasoning.
\begin{table*}[!htb]
\caption{The payoff matrices of all games.}
\centering
% Top row: Tables a and b
\begin{subtable}{0.19\linewidth}
\caption*{\#$1$: $2\times2$ and symmetric}
\centering
\renewcommand\arraystretch{0.8}   
\setlength{\tabcolsep}{0pt}      
\begin{tabular}{c cc}   
      & \textbf{A} & \textbf{B} \\ %\hline
\textbf{A~~}   & \TriCell{3}{3} & \TriCell{5}{0} \\ %\hline
\textbf{B~~}   & \TriCell{0}{5} & \TriCell{1}{1} \\
\end{tabular}
\label{tab:game_matrix_1}
\end{subtable}
\hfill
\begin{subtable}{0.19\linewidth}
\caption*{\#$2$: $2\times2$ and symmetric}
\centering
\renewcommand\arraystretch{0.8}  
\setlength{\tabcolsep}{0pt}      
\begin{tabular}{c cc}   
      & \textbf{A} & \textbf{B} \\ %\hline
\textbf{A~~}   & \TriCell{4}{4} & \TriCell{2}{1} \\ %\hline
\textbf{B~~}   & \TriCell{1}{2} & \TriCell{3}{3} \\
\end{tabular}
\label{tab:game_matrix_2}
\end{subtable}
\hfill
\begin{subtable}{0.19\linewidth}
\caption*{\#$3$: $2\times2$ and symmetric}
\centering
\renewcommand\arraystretch{0.8}  
\setlength{\tabcolsep}{0pt}      
\begin{tabular}{c cc}   
      & \textbf{A} & \textbf{B} \\ %\hline
\textbf{A~~}   & \TriCell{2}{2} & \TriCell{3}{1} \\ %\hline
\textbf{B~~}   & \TriCell{1}{3} & \TriCell{0}{0} \\
\end{tabular}
\label{tab:game_matrix_3}
\end{subtable}
\hfill
\begin{subtable}{0.19\linewidth}
\caption*{\#$4$: $2\times2$ and symmetric}
\centering
\renewcommand\arraystretch{0.8}  
\setlength{\tabcolsep}{0pt}      
\begin{tabular}{c cc}   
      & \textbf{A} & \textbf{B} \\ %\hline
\textbf{A~~}   & \TriCell{0}{0} & \TriCell{-3}{3} \\ %\hline
\textbf{B~~}   & \TriCell{3}{-3} & \TriCell{0}{0} \\
\end{tabular}
\label{tab:game_matrix_4}
\end{subtable}
\hfill
\begin{subtable}{0.19\linewidth}
\caption*{\#$5$: $2\times2$ and symmetric}
\centering
\renewcommand\arraystretch{0.8}  
\setlength{\tabcolsep}{0pt}      
\begin{tabular}{c cc}   
      & \textbf{A} & \textbf{B} \\ %\hline
\textbf{A~~}   & \TriCell{5}{5} & \TriCell{4}{3} \\ %\hline
\textbf{B~~}   & \TriCell{3}{4} & \TriCell{2}{2} \\
\end{tabular}
\label{tab:game_matrix_5}
\end{subtable}

\begin{subtable}{0.19\linewidth}
\caption*{\#$6$: $2\times2$ and symmetric}
\centering
\renewcommand\arraystretch{0.8}  
\setlength{\tabcolsep}{0pt}      
\begin{tabular}{c cc}   
      & \textbf{A} & \textbf{B} \\ %\hline
\textbf{A~~}   & \TriCell{-2}{-2} & \TriCell{0}{3} \\ %\hline
\textbf{B~~}   & \TriCell{3}{0} & \TriCell{1}{1} \\
\end{tabular}
\label{tab:game_matrix_6}
\end{subtable}
\hfill
\begin{subtable}{0.19\linewidth}
\caption*{\#$7$: $2\times2$ and symmetric}
\centering
\renewcommand\arraystretch{0.8}   
\setlength{\tabcolsep}{0pt}      
\begin{tabular}{c cc}   
      & \textbf{A} & \textbf{B} \\ %\hline
\textbf{A~~}   & \TriCell{1}{1} & \TriCell{2}{0} \\ %\hline
\textbf{B~~}   & \TriCell{0}{2} & \TriCell{3}{3} \\
\end{tabular}
\label{tab:game_matrix_7}
\end{subtable}
\hfill
\begin{subtable}{0.19\linewidth}
\caption*{\#$8$: $2\times2$ and asymmetric}
\centering
\renewcommand\arraystretch{0.8}  
\setlength{\tabcolsep}{0pt}       
\begin{tabular}{c cc}   
      & \textbf{A} & \textbf{B} \\ %\hline
\textbf{A~~}   & \TriCell{2}{3} & \TriCell{4}{1} \\ %\hline
\textbf{B~~}   & \TriCell{1}{2} & \TriCell{3}{4} \\
\end{tabular}
\label{tab:game_matrix_8}
\end{subtable}
\hfill
\begin{subtable}{0.19\linewidth}
\caption*{\#$9$: $2\times2$ and asymmetric}
\centering
\renewcommand\arraystretch{0.8}  
\setlength{\tabcolsep}{0pt}       
\begin{tabular}{c cc}   
      & \textbf{A} & \textbf{B} \\ %\hline
\textbf{A~~}   & \TriCell{4}{3} & \TriCell{5}{0} \\ %\hline
\textbf{B~~}   & \TriCell{3}{2} & \TriCell{2}{1} \\
\end{tabular}
\label{tab:game_matrix_9}
\end{subtable}
\hfill
\begin{subtable}{0.19\linewidth}
\caption*{\#$10$: $2\times2$ and asymmetric}
\centering
\renewcommand\arraystretch{0.8}   
\setlength{\tabcolsep}{0pt}       
\begin{tabular}{c cc}   
      & \textbf{A} & \textbf{B} \\ %\hline
\textbf{A~~}   & \TriCell{-3}{3} & \TriCell{5}{2} \\ %\hline
\textbf{B~~}   & \TriCell{1}{3} & \TriCell{0}{-2}
\end{tabular}
\label{tab:game_matrix_10}
\end{subtable}

\begin{subtable}{0.19\linewidth}
\caption*{\#$11$: $3\times3$ and symmetric}
\centering
\renewcommand\arraystretch{0.8}  
\setlength{\tabcolsep}{0pt}      
\scalebox{0.7}{
\begin{tabular}{cc cc}   
      & \textbf{A} & \textbf{B} &\textbf{C}\\ %\hline
\textbf{A~~}   & \TriCell{3}{3}	& \TriCell{5}{0} & \TriCell{1}{2} \\ %\hline
\textbf{B~~}   & \TriCell{0}{5}	& \TriCell{0}{0} & \TriCell{4}{1} \\
\textbf{C~~}   & \TriCell{2}{1}	& \TriCell{1}{4} & \TriCell{4}{4} \\
\end{tabular}}
\label{tab:game_matrix_11}
\end{subtable}
\hfill
\begin{subtable}{0.19\linewidth}
\caption*{\#$12$: $3\times3$ and symmetric}
\centering
\renewcommand\arraystretch{0.8}  
\setlength{\tabcolsep}{0pt}       
\scalebox{0.7}{
\begin{tabular}{cc cc}   
      & \textbf{A} & \textbf{B} &\textbf{C}\\ %\hline
\textbf{A~~}   & \TriCell{3}{3}	& \TriCell{5}{0} & \TriCell{1}{2} \\ %\hline
\textbf{B~~}   & \TriCell{0}{5}	& \TriCell{0}{0} & \TriCell{4}{1} \\
\textbf{C~~}   & \TriCell{2}{1}	& \TriCell{1}{4} & \TriCell{4}{4} \\
\end{tabular}}
\label{tab:game_matrix_12}
\end{subtable}
\hfill
\begin{subtable}{0.19\linewidth}
\caption*{\#$13$: $3\times3$ and symmetric}
\centering
\renewcommand\arraystretch{0.8}  
\setlength{\tabcolsep}{0pt}      
\scalebox{0.7}{
\begin{tabular}{cc cc}   
      & \textbf{A} & \textbf{B} &\textbf{C}\\ %\hline
\textbf{A~~}   & \TriCell{3}{3}	& \TriCell{5}{0} & \TriCell{4}{2} \\ %\hline
\textbf{B~~}   & \TriCell{0}{5}	& \TriCell{2}{2} & \TriCell{3}{1} \\
\textbf{C~~}   & \TriCell{2}{4}	& \TriCell{1}{3} & \TriCell{3}{3} \\
\end{tabular}}
\label{tab:game_matrix_13}
\end{subtable}
\hfill
\begin{subtable}{0.19\linewidth}
\caption*{\#$14$: $3\times3$ and asymmetric}
\centering
\renewcommand\arraystretch{0.8}   
\setlength{\tabcolsep}{0pt}       
\scalebox{0.7}{
\begin{tabular}{cc cc}   
      & \textbf{A} & \textbf{B} &\textbf{C}\\ %\hline
\textbf{A~~}   & \TriCell{2}{1}	& \TriCell{1}{0} & \TriCell{0}{-1} \\ %\hline
\textbf{B~~}   & \TriCell{1}{2}	& \TriCell{0}{0} & \TriCell{-1}{-2} \\
\textbf{C~~}   & \TriCell{1}{3}	& \TriCell{2}{1} & \TriCell{2}{-1} \\
\end{tabular}}
\label{tab:game_matrix_14}
\end{subtable}
\hfill
\begin{subtable}{0.19\linewidth}
\caption*{\#$15$: $3\times3$ and asymmetric}
\centering
\renewcommand\arraystretch{0.8}  
\setlength{\tabcolsep}{0pt}       
\scalebox{0.7}{
\begin{tabular}{cc cc}   
      & \textbf{A} & \textbf{B} &\textbf{C}\\ %\hline
\textbf{A~~}   & \TriCell{3}{5}	& \TriCell{4}{4} & \TriCell{5}{2} \\ %\hline
\textbf{B~~}   & \TriCell{4}{3}	& \TriCell{4}{4} & \TriCell{5}{1} \\
\textbf{C~~}   & \TriCell{5}{2}	& \TriCell{5}{1} & \TriCell{3}{3} \\
\end{tabular}}
\label{tab:game_matrix_15}
\end{subtable}
\label{tab:all_game_matrices_3}
\end{table*}

In the Memory Mechanism, we implement a dedicated \textbf{Historical Game Records} module that stores comprehensive interaction data from previous game-playing. 
This module systematically stores two key elements for each round of game-playing: 
(1) both agents' chosen actions, and (2) both agent's utilities gained based on the chosen actions. 
Here, the parameter \textbf{records} specifically refers to this structured compilation of historical behavioral patterns for both agents within the game-theoretic framework.
%\clearpage

\subsubsection*{Chat \& Memory Mechanism}
Under the Chat \& Memory Mechanism, there is three kinds of prompt templates, that is the Consultation prompt template of the agent (\Cref{fig:prompt_chat_memory_mechanism}), the Response prompt template of his advisor (\Cref{fig:prompt_chat_memory_friend_conversation}), and the Decision prompt template of the agent (\Cref{fig:prompt_chat_memory_choose}).
We integrate historical game-playing information into the prompts for the LLM, enabling two LLMs to conduct conversation with historical context. Two rounds of historically grounded interactions are incorporated into the prompt of the game-playing LLM. Under this mechanism, the LLM’s decision-making is jointly influenced by the Chat Mechanism and the Memory Mechanism. 

Figure \ref{fig:prompt_chat_memory_mechanism} represents the prompt for the LLM to generate conversation with its advisor. Figure \ref{fig:prompt_chat_memory_friend_conversation} represents the prompt for the LLM's companion to generate responses. Figure \ref{fig:prompt_chat_memory_choose} represents the prompt for the LLM's action choice after completing the conversation. \textbf{conversation\_1} represents the conversations information between the LLM and its companion after 10 rounds of game-playing. \textbf{conversation\_2} represents the conversations information between the LLM and its companion after 20 rounds of game-playing.

\subsection{Payoff Matrices of Normal-form Games}
\label{sec:game_matrix_all}
In this section, we present the payoff matrices of $15$ normal-form games that are used to construct the training dataset. 
Among these games, Games \#$1$ to \#$7$ and \#$11$ to \#$13$ are symmetric games, while the remaining are asymmetric games. 
In each symmetric game, both players follow the same game rule and have identical payoff structures. 
In each asymmetric game, payoffs are unevenly distributed, and both players face different decision challenges based on their unique roles. 
\begin{table*}[!htb]
\centering
	\caption{Mean and variance of the predicted level distribution in the Level-K CH model when $\hat{k}=3$. Red cells indicate a negative effect and green cells indicate a positive effect. The bold cells represent the mechanisms with the best LLM strategy reasoning capability.}
	\resizebox{1\textwidth}{!}{
	\begin{tabular}{c|cc|cc|cc|cc}

	\multirow{2}{*}{\diagbox{\textbf{Model}}{\textbf{Mechanisms}}}		& \multicolumn{2}{c|}{\textbf{Baseline}}&\multicolumn{2}{c|}{\textbf{Chat Mechanism}}	&\multicolumn{2}{c|}{\textbf{Memory Mechanism}}&\multicolumn{2}{c}{\textbf{Chat \& Memory}}

	\\
		&\textbf{Var}	&\textbf{Mean}	&\textbf{Var}	&\textbf{Mean}	&\textbf{Var}   &\textbf{Mean}
		&\textbf{Var}  &\textbf{Mean}\\
		\Xhline{1pt}
Deepseek-chat&0.0084&0.4438&0.0068&\cellcolor[RGB]{175, 240, 175}0.7471&0.0081&\cellcolor[RGB]{175, 240, 175}0.8668&0.0023&\cellcolor[RGB]{175, 240, 175}\textbf{0.9054}\\
Gemini-1.5-pro&0.0078&1.0014&0.0184&\cellcolor[RGB]{240, 175, 175}0.5713&0.0032&\cellcolor[RGB]{175, 240, 175}\textbf{1.0736}&0.0011&\cellcolor[RGB]{240, 175, 175}0.4756\\
Gpt-4o-mini&0.0141&0.5663&0.0068&\cellcolor[RGB]{240, 175, 175}0.1564&0.0343&\cellcolor[RGB]{175, 240, 175}\textbf{0.8817}&0.0400&\cellcolor[RGB]{175, 240, 175}0.7411\\
Llama-3.1-70b&0.0055&0.6690&0.0052&\cellcolor[RGB]{240, 175, 175}0.4488&0.0092&\cellcolor[RGB]{175, 240, 175}\textbf{0.9086}&0.0070&\cellcolor[RGB]{240, 175, 175}0.6090\\
Qwen-max&0.02190&0.5020&0.0018&\cellcolor[RGB]{240, 175, 175}0.2007&0.0210&\cellcolor[RGB]{175, 240, 175}\textbf{0.8950}&0.01400&\cellcolor[RGB]{240, 175, 175}0.4755\\
Gpt-4o&0.0087&1.1022&0.0055&\cellcolor[RGB]{240, 175, 175}0.6025&0.0012&\cellcolor[RGB]{175, 240, 175}\textbf{1.1094}&0.0204&\cellcolor[RGB]{240, 175, 175}0.9715\\
		\hline
	\end{tabular}
	}
	\label{tab:lk_var_mean_k3}
\end{table*}
\begin{table*}[!htb]
\centering
	\caption{Mean and variance of the predicted level distribution in the Poisson CH model when $\hat{k}=3$. Red cells indicate a negative effect and green cells indicate a positive effect. The bold cells represent the mechanisms with the best LLM strategy reasoning capability.}
	\resizebox{1\textwidth}{!}{
	\begin{tabular}{c|cc|cc|cc|cc}

	\multirow{2}{*}{\diagbox{\textbf{Model}}{\textbf{Mechanisms}}}	& \multicolumn{2}{c|}{\textbf{Baseline}}&\multicolumn{2}{c|}{\textbf{Chat Mechanism}}	&\multicolumn{2}{c|}{\textbf{Memory Mechanism}}&\multicolumn{2}{c}{\textbf{Chat \& Memory}}
	\\
		&\textbf{Var}	&\textbf{Mean}	&\textbf{Var}	&\textbf{Mean}	&\textbf{Var}   &\textbf{Mean}
		&\textbf{Var}  &\textbf{Mean}\\
		\Xhline{1pt}
Deepseek-chat&0.0006&0.3628&0.0005&\cellcolor[RGB]{175, 240, 175}0.5856&0.0002&\cellcolor[RGB]{175, 240, 175}\textbf{0.5980}&0.0001&\cellcolor[RGB]{175, 240, 175}0.5782\\
Gemini-1.5-pro&0.0007&0.6447&0.0122&\cellcolor[RGB]{240, 175, 175}0.2714&0.0001&\cellcolor[RGB]{175, 240, 175}\textbf{0.6760}&0.0019&\cellcolor[RGB]{240, 175, 175}0.3025\\
Gpt-4o-mini&0.0022&0.2566&0.0022&\cellcolor[RGB]{240, 175, 175}0.0473&0.0009&\cellcolor[RGB]{175, 240, 175}\textbf{0.6155}&0.0096&\cellcolor[RGB]{175, 240, 175}0.4177\\
Llama-3.1-70b&0.0035&0.4568&0.0025&\cellcolor[RGB]{240, 175, 175}0.4117&0.0000&\cellcolor[RGB]{175, 240, 175}\textbf{0.5734}&0.0009&\cellcolor[RGB]{240, 175, 175}0.3661\\
Qwen-max&0.0069&0.2034&0.0013&\cellcolor[RGB]{240, 175, 175}0.1634&0.0029&\cellcolor[RGB]{175, 240, 175}\textbf{0.6423}&0.0039&\cellcolor[RGB]{175, 240, 175}0.3728\\
Gpt-4o&0.0001&0.6442&0.0057&\cellcolor[RGB]{240, 175, 175}0.3935&0.0006&\cellcolor[RGB]{175, 240, 175}\textbf{0.6739}&0.0070&\cellcolor[RGB]{175, 240, 175}0.6542\\
		\hline
	\end{tabular}
	}
	\label{tab:p_ch_var_mean_k3}
\end{table*}
\begin{table*}[!htb]
\centering
	\caption{Mean and variance of the predicted level distribution in the Level-K CH model when $\hat{k}=5$. Red cells indicate a negative effect and green cells indicate a positive effect. The bold cells represent the mechanisms with the best LLM strategy reasoning capability.}
	\resizebox{1\textwidth}{!}{
	\begin{tabular}{c|cc|cc|cc|cc}

	\multirow{2}{*}{\diagbox{\textbf{Model}}{\textbf{Mechanisms}}}		& \multicolumn{2}{c|}{\textbf{Baseline}}&\multicolumn{2}{c|}{\textbf{Chat Mechanism}}	&\multicolumn{2}{c|}{\textbf{Memory Mechanism}}&\multicolumn{2}{c}{\textbf{Chat \& Memory}}\\
		&\textbf{Var}	&\textbf{Mean}	&\textbf{Var}	&\textbf{Mean}	&\textbf{Var}   &\textbf{Mean}
		&\textbf{Var}  &\textbf{Mean}\\
		\Xhline{1pt}
Deepseek-chat&0.2252&0.9497&0.3766&\cellcolor[RGB]{175, 240, 175}1.3802&0.0645&\cellcolor[RGB]{175, 240, 175}\textbf{2.1454}&0.0595&\cellcolor[RGB]{175, 240, 175}1.2641\\
Gemini-1.5-prot&0.0074&1.0229&0.0085&\cellcolor[RGB]{240, 175, 175}0.6235&0.0880&\cellcolor[RGB]{175, 240, 175}\textbf{1.5287}&0.0196&\cellcolor[RGB]{240, 175, 175}0.7279\\
Gpt-4o-mini&0.0175&0.7996&0.0056&\cellcolor[RGB]{240, 175, 175}0.3505&0.0877&\cellcolor[RGB]{175, 240, 175}\textbf{1.2037}&0.0701&\cellcolor[RGB]{240, 175, 175}0.7690\\
Llama-3.1-70b&0.0970&1.5647&0.0766&\cellcolor[RGB]{240, 175, 175}0.8733&0.0481&\cellcolor[RGB]{175, 240, 175}\textbf{1.9442}&0.0133&\cellcolor[RGB]{240, 175, 175}1.3099\\
Qwen-max&0.0099&0.7341&0.0191&\cellcolor[RGB]{175, 240, 175}1.1367&0.0968&\cellcolor[RGB]{175, 240, 175}\textbf{1.6811}&0.0088&\cellcolor[RGB]{240, 175, 175}0.5641\\
Gpt-4o&0.1793&1.5321&0.2552&\cellcolor[RGB]{240, 175, 175}0.7852&0.0118&\cellcolor[RGB]{175, 240, 175}\textbf{1.8207}&0.0660&\cellcolor[RGB]{240, 175, 175}1.5123\\
		\hline
	\end{tabular}
	}
	\label{tab:lk_var_mean_k5}
\end{table*}
\begin{table*}[!htb]
\centering
	\caption{Mean and variance of the predicted level distribution in the Poisson CH model when $\hat{k}=5$. Red cells indicate a negative effect and green cells indicate a positive effect. The bold cells represent the mechanisms with the best LLM strategy reasoning capability.}
	\resizebox{1\textwidth}{!}{
	\begin{tabular}{c|cc|cc|cc|cc}

	\multirow{2}{*}{\diagbox{\textbf{Model}}{\textbf{Mechanisms}}}	& \multicolumn{2}{c|}{\textbf{Baseline}}&\multicolumn{2}{c|}{\textbf{Chat Mechanism}}	&\multicolumn{2}{c|}{\textbf{Memory Mechanism}}&\multicolumn{2}{c}{\textbf{Chat \& Memory}}
	\\
		&\textbf{Var}	&\textbf{Mean}	&\textbf{Var}	&\textbf{Mean}	&\textbf{Var}   &\textbf{Mean}
		&\textbf{Var}  &\textbf{Mean}\\
		\Xhline{1pt}
Deepseek-chat&0.0064&0.8965&0.0156&\cellcolor[RGB]{175, 240, 175}1.4764&0.0107&\cellcolor[RGB]{175, 240, 175}\textbf{1.7085}&0.0123&\cellcolor[RGB]{175, 240, 175}1.3906\\
Gemini-1.5-pro&0.0141&1.8390&0.1163&\cellcolor[RGB]{240, 175, 175}0.5820&0.0027&\cellcolor[RGB]{175, 240, 175}\textbf{2.0163}&0.0107&\cellcolor[RGB]{240, 175, 175}0.4546\\
Gpt-4o-mini&0.0000&0.8388&0.0028&\cellcolor[RGB]{240, 175, 175}0.0515&0.1038&\cellcolor[RGB]{175, 240, 175}\textbf{1.7262}&0.0912&\cellcolor[RGB]{175, 240, 175}0.8455\\
Llama-3.1-70b&0.0025&0.9147&0.0119&\cellcolor[RGB]{240, 175, 175}0.7845&0.0032&\cellcolor[RGB]{175, 240, 175}\textbf{1.5387}&0.0254&\cellcolor[RGB]{175, 240, 175}1.0100\\
Qwen-max&0.0954&0.6266&0.0030&\cellcolor[RGB]{240, 175, 175}0.2003&0.0811&\cellcolor[RGB]{175, 240, 175}\textbf{1.8581}&0.0474&\cellcolor[RGB]{175, 240, 175}0.7365\\
Gpt-4o&0.0060&1.8965&0.0083&\cellcolor[RGB]{240, 175, 175}0.8311&0.0073&\cellcolor[RGB]{175, 240, 175}\textbf{2.0388}&0.1502&\cellcolor[RGB]{240, 175, 175}1.8687\\
		\hline
	\end{tabular}
	}
	\label{tab:p_ch_var_mean_k5}
\end{table*}

\begin{table*}[!htb]
 \centering
  	\caption{Average utility under Baseline.}
 	\resizebox{1\textwidth}{!}{
 	\begin{tabular}{c|c|c|c|c|c|c}
		\multirow{2}{*}{\diagbox{\textbf{Model}}{\textbf{Opponent}}}
			&&&&&&		\\
 	&\textbf{DeepSeek-chat}	&\textbf{Gemini-1.5-pro}	&\textbf{GPT-4o-mini}	&\textbf{Llama-3.1-70b}	&\textbf{Qwen-max}   &\textbf{GPT-4o}\\
 		\Xhline{1pt}
 DeepSeek-chat&1.9733&1.7483&2.6450&1.7117&2.6150&2.3300\\
 Gemini-1.5-pro&2.0367&1.9167&2.6367&1.7833&2.7033&2.4833\\
 GPT-4o-mini&1.7600&1.7900&2.5883&1.6117&2.6650&2.2900\\
 Llama-3.1-70b&1.7483&1.6583&2.5533&1.7600&2.4083&2.2433\\
 Qwen-max&1.6650&1.7850&2.5200&1.5083&2.6350&2.1917\\
 GPT-4o&1.9550&1.9717&2.6583&1.7783&2.7000&2.5467 \\
 		\hline
 	\end{tabular}
 	}
 	\label{tab:LLM_utility_baseline}
 \end{table*}
 
  \begin{table*}[!htb]
 \centering
  	\caption{Average utility under the Chat Mechanism.}
 	\resizebox{1\textwidth}{!}{
 	\begin{tabular}{c|c|c|c|c|c|c}
		\multirow{2}{*}{\diagbox{\textbf{Model}}{\textbf{Opponent}}}
			&&&&&& 		\\
 	&\textbf{DeepSeek-chat}	&\textbf{Gemini-1.5-pro}	&\textbf{GPT-4o-mini}	&\textbf{Llama-3.1-70b}	&\textbf{Qwen-max}   &\textbf{GPT-4o}\\
 		\Xhline{1pt}
Deepseek-chat&2.6433&1.8383&2.6717&2.0200&2.7183&2.4700\\
Gemini-1.5-pro&2.3100&2.2267&2.0967&2.0633&2.6767&2.3200\\
Gpt-4o-mini&2.3367&2.4517&2.7900&2.2083&2.4133&2.1833\\
Llama-3.1-70b&2.5500&2.5650&2.7267&1.9767&2.8117&2.1767\\
Qwen-max&2.4333&1.9233&2.8117&2.2017&2.7067&2.1133\\
Gpt-4o&2.4433&2.3083&2.9367&2.2150&2.8183&2.5867\\
 		\hline
 	\end{tabular}
 	}
 	\label{tab:LLM_utility_chat}
 \end{table*}
  \begin{table*}[!htb]
 \centering
  	\caption{Average utility under the Memory Mechanism.}
 	\resizebox{1\textwidth}{!}{
 	\begin{tabular}{c|c|c|c|c|c|c}
		\multirow{2}{*}{\diagbox{\textbf{Model}}{\textbf{Opponent}}}
			&&&&&& 		\\
 	&\textbf{DeepSeek-chat}	&\textbf{Gemini-1.5-pro}	&\textbf{GPT-4o-mini}	&\textbf{Llama-3.1-70b}	&\textbf{Qwen-max}   &\textbf{GPT-4o}\\
 		\Xhline{1pt}
Deepseek-chat&2.7100&2.4467&2.8133&2.2900&2.6067&2.3750\\
Gemini-1.5-pro&2.4700&2.6367&2.9550&2.0950&2.5367&2.4183\\
Gpt-4o-mini&2.7150&2.3133&2.8400&2.1550&2.6133&2.4550\\
Llama-3.1-70b&2.3450&2.0333&2.6267&1.9533&2.1983&2.0200\\
Qwen-max&2.6217&2.4800&2.8933&2.1683&2.7117&2.5067\\
Gpt-4o&2.6467&2.4967&2.8517&2.1300&2.6317&2.3583\\
 		\hline
 	\end{tabular}
 	}
 	\label{tab:LLM_utility_memory}
 \end{table*}
   \begin{table*}[!htb]
 \centering
  	\caption{Average utility under the Chat \& Memory Mechanism.}
 	\resizebox{1\textwidth}{!}{
 	\begin{tabular}{c|c|c|c|c|c|c}
		\multirow{2}{*}{\diagbox{\textbf{Model}}{\textbf{Opponent}}}
			&&&&&& 		\\
 	&\textbf{DeepSeek-chat}	&\textbf{Gemini-1.5-pro}	&\textbf{GPT-4o-mini}	&\textbf{Llama-3.1-70b}	&\textbf{Qwen-max}   &\textbf{GPT-4o}\\
 		\Xhline{1pt}
Deepseek-chat&2.3733&2.4517&2.9617&2.1167&2.5367&2.3117\\
Gemini-1.5-pro&2.1850&2.3517&2.4617&1.9500&2.2750&2.4033\\
Gpt-4o-mini&2.3933&2.0883&2.3067&1.9283&2.0683&1.9117\\
Llama-3.1-70b&2.3583&1.9667&2.2900&2.0767&2.2767&1.8100\\
Qwen-max&2.2933&2.2100&2.4033&2.2667&2.3233&2.0750\\
Gpt-4o&2.7717&2.3050&2.8350&1.9967&2.2517&2.7267\\
 		\hline
 	\end{tabular}
 	}
 	\label{tab:LLM_utility_chat_memory}
 \end{table*}
   \begin{table*}[!htb]
 \centering
  	\caption{Utility variance under different mechanisms.}
 	\resizebox{1\textwidth}{!}{
 	\begin{tabular}{c|c|c|c|c}
 	\multirow{2}{*}{\diagbox{\textbf{Model}}{\textbf{Mechanism}}}
			&&&& 		\\
 	&\textbf{Baseline}	&\textbf{Chat Mechanism}	&\textbf{Memory Mechanism}	&\textbf{Chat \& Memory}\\
 		\Xhline{1pt}
Deepseek-chat&0.1752&0.1398&0.0413&0.0810\\
Gemini-1.5-pro&0.1566&0.0486&0.0793&0.0342\\
Gpt-4o-mini&0.2084&0.0485&0.0658&0.0387\\
Llama-3.1-70b&0.1494&0.1055&0.0648&0.0462\\
Qwen-max&0.2190&0.1212&0.0601&0.0125\\
Gpt-4o&0.1684&0.0810&0.0635&0.1177\\
 		\hline
 	\end{tabular}
 	}
 	\label{tab:LLM_utility_variance}
 \end{table*}
In each $2\times 2$ game, both players have two possible actions to choose from. In each $3\times 3$ game, both players have three possible actions to choose from. Among these games, Games \#$1$ to \#$10$ are $2\times 2$ games, while the remaining are $3\times 3$ games.
In each payoff matrix, there is one row player and one column player, where each player has $2$ or $3$ possible actions to choose from. 
For any given combination of actions chosen by the two players, the ordered pair $(A_r, B_c)$ represents their respective payoffs, with the first element $A_r$ denoting the row player's payoff and the second element $B_c$ denoting the column player's payoff. 
%\clearpage
\subsection{Predicted Level Distributions under Different Maximum Level $\hat{k}$}
\label{sec:appendix_exp_lk}
In this section, we report the mean $k$ for $\hat{k}=3$ and $\hat{k}=5$, along with the variance of $k$ for LLMs interacting with different opponents.

The experimental results exhibit the same trend for $\hat{k}=3$ and $\hat{k}=4$.
In the experimental results with $\hat{k}=5$, GPT-4o exhibits a variance of 0.1502 under Chat \& Memory, indicating redundancy at $\hat{k}=5$. A more suitable choice is $\hat{k}=4$.  
 
 As introduced in Section \ref{sec:exp}, the action choice strategy of the LLM remains consistent as $\hat{k}$ increases. Under different $\hat{k}$ values, the probabilities of $k=4$ and $k=3$ influence the mean $k$, which explains why the mean $k$ varies across different $\hat{k}$ values. The difference in mean $k$ across $\hat{k}$ values cannot be used to evaluate the strategic reasoning capability of different LLMs. 
%\clearpage

\subsection{Utility Performance as Evaluation Metric}
\label{sec:appendix_exp_u}
In this section, we present the average utility of the LLM when the opponent varies. The average utility of LLMs exhibits substantial variance. The Memory Mechanism stabilizes this utility, whereas its variance remains higher than the Cognitive Hierarchy models.
 
As shown in Tables \ref{tab:LLM_utility_baseline}, \ref{tab:LLM_utility_chat}, \ref{tab:LLM_utility_memory}, and \ref{tab:LLM_utility_chat_memory}, the average utility of LLMs varies significantly when playing games against different opponents. The utility cannot evaluate the strategic reasoning capability of LLMs. Table \ref{tab:LLM_utility_variance} presents the variance of LLM utilities across different opponents, where higher values indicate greater fluctuations in utility. The reduced variance with the Memory Mechanism demonstrates improved strategic reasoning capability of LLMs after its incorporation. This experimental result is more pronounced in Cognitive Hierarchy models.

\label{sec:appendix}

\end{document}